\documentclass[letterpaper, 10 pt, conference]{ieeeconf}
\IEEEoverridecommandlockouts

\usepackage[utf8]{inputenc} 
\usepackage[T1]{fontenc}
\usepackage{amsmath}
\usepackage{amsfonts}
\usepackage{float}
\usepackage{graphicx}
\usepackage{booktabs}
\usepackage[font=small]{caption}
\usepackage{subcaption}
\usepackage[ruled]{algorithm2e}
\usepackage{booktabs} 
\usepackage{enumerate}
\usepackage{times}
\usepackage{soul}
\usepackage{url}
\usepackage{hyperref}
\usepackage{bm}
\usepackage{adjustbox}
\usepackage{xcolor}
\usepackage{placeins}
\usepackage[utf8]{inputenc}
\usepackage{booktabs}
\usepackage{array, multirow}
\usepackage{cleveref}
\usepackage{tikz}
\usepackage{wrapfig}
\usepackage{adjustbox}
\usepackage{multirow}

\newcommand{\bb}{\mathbf}  
\begin{document}

\newcommand\mycommfont[1]{\footnotesize\rmfamily\textcolor{blue}{#1}}
\usetikzlibrary{arrows.meta}
\usetikzlibrary{positioning}
\tikzstyle{decision} = [diamond, draw, fill=blue!20, 
    text width=6em, text badly centered, node distance=3cm, inner sep=0pt]
\tikzstyle{block} = [rectangle, draw, fill=gray!10, 
    text width=10em, very thick, text centered, rounded corners, minimum height=2.2em]
\tikzstyle{line} = [draw, -{latex[scale=15.0]}]
\tikzstyle{cloud} = [draw, ellipse,fill=red!20, node distance=3cm,
    minimum height=2em]
\setlength{\fboxrule}{1pt}
\setlength{\fboxsep}{0pt}  

\setlength{\abovecaptionskip}{2pt}
\setlength{\belowcaptionskip}{2pt}

\title{\LARGE \bf Instructing Robots by Sketching: Learning from Demonstration via Probabilistic Diagrammatic Teaching
}

\author{Weiming Zhi$^{1,*}$ \and Tianyi Zhang$^{1}$ \and Matthew Johnson-Roberson$^{1}$
\thanks{$^{*}$email: {\tt\small wzhi@andrew.cmu.edu}.}%
\thanks{$^{1}$ Robotics Institute, Carnegie Mellon University, Pittsburgh, PA, USA}%
}

\maketitle

\begin{abstract}
Learning from Demonstration (LfD) enables robots to acquire new skills by imitating expert demonstrations, allowing users to communicate their instructions intuitively. Recent progress in LfD often relies on kinesthetic teaching or teleoperation as the medium for users to specify the demonstrations. Kinesthetic teaching requires physical handling of the robot, while teleoperation demands proficiency with additional hardware. This paper introduces an alternative paradigm for LfD called \emph{Diagrammatic Teaching}. Diagrammatic Teaching aims to teach robots novel skills by prompting the user to sketch out demonstration trajectories on 2D images of the scene, these are then synthesised as a generative model of motion trajectories in 3D task space. Additionally, we present the Ray-tracing Probabilistic Trajectory Learning (RPTL) framework for Diagrammatic Teaching. RPTL extracts time-varying probability densities from the 2D sketches, then applies ray-tracing to find corresponding regions in 3D Cartesian space, and fits a probabilistic model of motion trajectories to these regions. New motion trajectories, which mimic those sketched by the user, can then be generated from the probabilistic model. We empirically validate our framework both in simulation and on real robots, which include a fixed-base manipulator and a quadruped-mounted manipulator.
\end{abstract}


\section{Introduction}
Learning from Demonstration (LfD) enables robots to learn novel motions by mimicking a collected set of expert trajectories \cite{ilbook}. LfD is particularly appealing in its ability to specify complex robot movements in the absence of explicit programming or cost design, thereby empowering non-roboticists to teach a robot how to act. Demonstrations are typically collected via kinesthetic teaching where a human physically handles the robot, or via teleoperation where the expert uses a remote controller to collect demonstrations. These approaches can be limiting, as they may require co-location with the robot or proficiency with specialised hardware. These challenges are further amplified when attempting to provide demonstrations to mobile manipulators.

  

\begin{figure}
  \includegraphics[width=\linewidth]{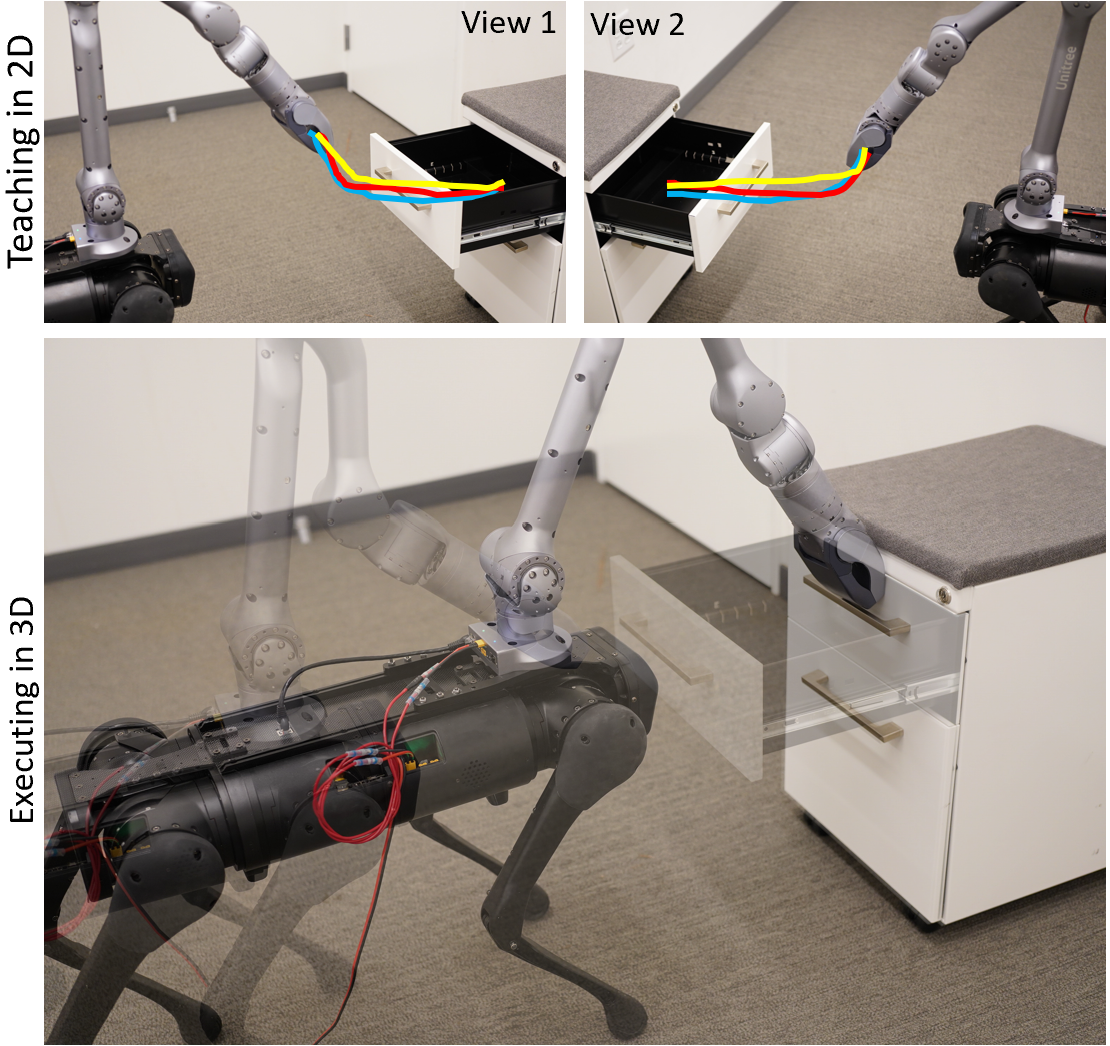}
  \caption{\emph{Diagrammatically teaching} a quadruped with a mounted arm to shut a drawer, by sketching robot demonstrations over 2D images. The user sketches trajectories of the desired movement over images of the scene (\underline{top left and right}). The robot then learns to execute the skill and close the drawer (\underline{bottom}).}
  \label{fig:draw_close_example}
\end{figure}

This paper proposes \emph{Diagrammatic Teaching}, as a paradigm for LfD where a small number of demonstrations are provided by the user through sketches over static two-dimensional images of the environment scene from different views. These images can either be captured by a camera or generated from a scene representation, such as a NeRF model \cite{mildenhall2020nerf}. Diagrammatic Teaching seeks to enable the teaching of new skills to the robot from the sketched demonstrations. We note that humans have a remarkable ability to infer instructions from crude diagrammatic sketches, and we wish to endow robots with this same capability. 

Correspondingly, we present \emph{Ray-tracing Probabilistic Trajectory Learning} (RPTL), a framework for Diagrammatic Teaching. RPTL extracts time-varying probability distributions from user-provided sketches over 2D images, and then uses ray tracing to fit a probabilistic distribution of continuous motion trajectories in the 3D task space of a robot. This can be achieved with sketched demonstrations on as few as two images taken at different view poses. We can then adapt the trained model to generate trajectories that account for novel starting positions. We present extensive evaluations of RPTL in simulation and on real-world robots.   

Concretely, the technical contributions of this paper include:
\begin{enumerate}
\item The formulation of Diagrammatic Teaching Problem, an alternative set-up for LfD where the user provides demonstrations of motion via sketching, avoiding the need for physical interaction with the robot;
\item Ray-tracing Probabilistic Trajectory Learning (RPTL), a novel framework for Diagrammatic Teaching, which estimates probabilistic distributions of task space motion trajectories from 2D sketches. 
\end{enumerate}

In the subsequent sections, we first briefly review the related work (\cref{sec:Related_Work}) and introduce Diagrammatic Teaching as a paradigm for LfD (\cref{sec:diagrammatic_teaching}). Then, we introduce and elaborate on Ray-tracing Probabilistic Trajectory Learning (RPTL) (\cref{sec:RPTL}). Finally, we present empirical evaluations of RPTL, before concluding and presenting future research directions (\cref{sec:conclusions}). 

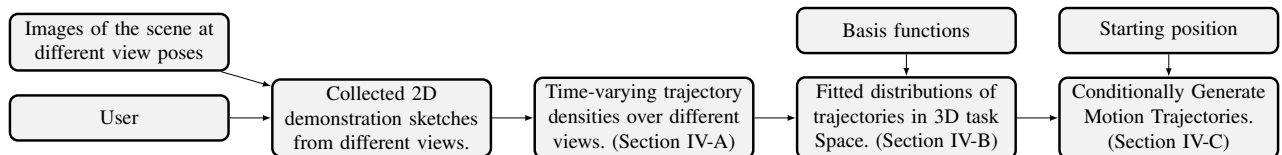
\begin{figure*}[t]
\begin{adjustbox}{width=0.95\textwidth,center} 
\begin{tikzpicture}[node distance = 1.2cm]
    \node [block] (Expert){User};
    \node [block, above of=Expert, node distance=1.3cm] (Images){Images of the scene at different view poses};
    
    \node [block,right of=Expert, node distance=4.5cm] (Demonstrations){Collected 2D demonstration sketches from different views.};
    \node [block,right of=Demonstrations, node distance=4.5cm] (Densities){Time-varying trajectory densities over different views. (\Cref{subsec:view_space})};
    \node [block,right of=Densities, node distance=4.5cm] (Generative){Fitted distributions of trajectories in 3D task Space. (\Cref{subsec:trajectory_fitting})};
    \node [block,above of=Generative, node distance=1.5cm] (Basis){Basis functions};
    \node [block,right of=Generative, node distance=4.5cm] (condition){Conditionally Generate Motion Trajectories. (\Cref{subsec:trajector_gen})};
    \node [block,above of=condition, node distance=1.5cm] (position){Starting position};
    

    \path [line] (Images) -- (Demonstrations);
    \path [line] (Expert) -- (Demonstrations);
    \path [line] (Demonstrations) -- (Densities);
    \path [line] (Densities) -- (Generative);
    \path [line] (Basis) -- (Generative);
    \path [line] (Generative) -- (condition);
    \path [line] (position) -- (condition);
\end{tikzpicture}
\end{adjustbox}
\caption{Overview and components of Ray-tracing Probabilistic Trajectory Learning for Diagrammatic Teaching.}\label{fig:Overview}
\end{figure*}

\section{Related Work}\label{sec:Related_Work}
\textbf{Learning for Demonstration:} Learning for Demonstration (LfD) is broadly a strategy to teach robots novel skills by learning on a small set of expert demonstrations. LfD circumvents the explicit programming of actions or the hand-designing of planning costs, which are time-consuming and require expertise \cite{ravichandar2020recent, ilbook}. Many attempts at LfD collect demonstrations by \emph{Kinesthetic Teaching}, where the user physically handles the robot to show it the desired motion while recording the robot's joint or end-effector coordinates. These include Dynamical Movement Primitive (DMP) approaches \cite{learning_mps,Ijspeert2013DynamicalMP}, Probabilistic Movement Primitives \cite{promp}, and stable dynamical system approaches \cite{SEDS,SEDS2,Diff_templates}. In these methods, demonstrations can be difficult to obtain as they requires the user to physically interact with the robot. Another approach for obtaining demonstrations is teleoperation, where a human provides demonstrations via a remote controller. This enables humans to provide demonstrations even when not co-located with the robot, allowing larger demonstration datasets to be collected \cite{Mandlekar2018ROBOTURKAC}. However, collecting trajectories via teleoperation is non-trivial and requires proficiency with the remote controller. The use of virtual reality \cite{Zhang2017DeepIL} can also simplify the demonstration collection procedure but requires specialised hardware. Compared with these approaches, the proposed Diagrammatic Teaching interface is distinct in enabling sketching as a medium for users to specify demonstrations. 

\textbf{Ray Tracing for Neural Rendering:} Neural rendering methods trace rays from 2D images into 3D space \cite{nerf_review_frank}, allowing a 3D scene model to be built from a set of images captured at multiple views. This model can then be used to generate images of the scene from arbitrary views. NeRF models \cite{mildenhall2020nerf} have been the most prominent of neural volume rendering methods where neural networks are used to regress onto the density and colour of the scene. Many follow-up variants focus on speeding up NeRFs, including Decomposed Neural Fields \cite{Rebain2020DeRFDR}, Neural Sparse Voxel Fields \cite{liu2020fast_nerf}, and Instant Neural Graphics Primitives \cite{mueller2022instant}. Like NeRF methods, our proposed Ray-tracing Probabilistic Trajectory Learning (RPTL) framework also applies ray-tracing to learn a spatially 3D model from 2D data. Additionally, RPTL requires demonstrators to sketch onto images from separate views --- these can be either taken from cameras or generated from NeRF models.

\section{The Diagrammatic Teaching Problem}\label{sec:diagrammatic_teaching}

Diagrammatic teaching follows the typical setup for Learning for Demonstration: we are assumed to have a small dataset of demonstrated trajectories and seek to learn a generative model that produces trajectories to imitate the demonstrations. Diagrammatic Teaching is unique in that the trajectories provided are \textbf{not} trajectories in the robot's Cartesian end-effector space, nor its configuration space, but are instead 2D trajectories sketched onto images, while the desired generative model produces end-effector trajectories. 

Formally, we have a dataset $\mathcal{D}=\{(v^{j},\bm{\zeta}_{i}^{j})_{i=1}^{n_{j}}\}_{j=1}^{2}$, where $v^{1}$ and $v^{2}$ denote two unique views from which the demonstrations are collected, and $\bm{\zeta}_{i}^{j}$ denotes the $i^{th}$ trajectory collected from the $j^{th}$ view. The user is shown images of the scene rendered from the views, and the user sketches how the end-effector movement is expected to look from the views. Images from different views may be captured by cameras at different locations or rendered from 3D scene representations such as NeRF. The collected \emph{view space} trajectories $\bm{\zeta}$ contain sequences of normalised pixel coordinates $(x,y)$ along with a normalised time $t$, i.e. $\bm{\zeta}=\{{t}_{k},x_{k},y_{k}\}_{k=1}^{l}$. The length of the trajectory is denoted by $l$. We assume that $x,y,t$ are all normalised to be in $[0,1]$. 

We aim to learn the generative model of trajectories $p(\bm{\xi}|\mathcal{D})$, where $\bm{\xi}$ denotes motion trajectories in the robot's end-effector task space. Here, $\bm{\xi}$ are represented as functions that map from normalised time $t\in[0,1]$ to Cartesian space $\bb{x}\in\mathbb{R}^{3}$. \Cref{fig:draw_close_example} shows an example of applying Diagrammatic Teaching to teach a quadruped robot with a mounted manipulator to shut a drawer by simply providing it demonstrations as sketches in two different views. 

\section{Ray-tracing Probabilistic Trajectory Learning}\label{sec:RPTL}
In this section, we propose Ray-tracing Probabilistic Trajectory Learning (RPTL), a framework to solve the Diagrammatic Teaching problem. The overview of the different components in RPTL is outlined in \cref{fig:Overview}. The user is prompted to sketch trajectories onto images of the scene, which could be generated from a NeRF model or taken by cameras in multiple poses. The camera poses can be accurately estimated by tools such as COLMAP \cite{schoenberger2016sfm,schoenberger2016mvs}, or with identifiable shared objects, such as an AprilTag \cite{AprilTag}, in the images taken. RPTL constructs time-varying density estimates over the collected 2D trajectories and then uses ray-tracing to find regions in 3D Cartesian space corresponding to the densities. Subsequently, a distribution of continuous-time motion trajectories can be fitted over these regions, and end-effector motion trajectories can be generated from the distribution. 

\subsection{Density Estimation in View Space}\label{subsec:view_space}
We begin by estimating the time-varying densities over 2D coordinates of demonstrations from each view, which we denote as $p(x^{j},y^{j}|t)$ for $j\in\{1,2\}$. We use flexible normalizing flow models \cite{normalising_flow} to estimate the joint distribution $p^{j}(t,x^{j},y^{j})$, and assume that $p(t)$ is uniform over $[0,1]$. A normalizing flow model uses neural networks to learn an invertible and differentiable function that transforms the arbitrarily complex data distribution into a known base distribution, typically a standard Gaussian, $\mathcal{N}(0,\mathbf{I})$. Let $\bb{y}^{j}=[t,x^{j},y^{j}]$ be time-stamped coordinates from demonstrated trajectories from the $j^{th}$ view and $\bb{\hat{y}}^{j}\in\mathbb{R}^{3}$ be corresponding latent variables linked by invertible functions $g^{j}$, such that $\bb{\hat{y}}^{j}=g^{j}(\bb{y}^{j})$. The densities over $\bb{y}^{j}$ and $\bb{\hat{y}}^{j}$ are linked by the change of variables, 
\begin{align}
    p(\bb{y}^{j})&=p(g^{j}(\bb{y}^{j}))\lvert\mathrm{det}\bb{J}^{j}(\bb{y}^{j})\lvert, \label{eqn:flow}\\
    p(\bb{\hat{y}}^{j})&=p({g^{j}}^{-1}(\bb{\hat{y}}^{j}))\lvert\mathrm{det}\bb{J}^{j}({g^{j}}^{-1}(\bb{\hat{y}}^{j}))\lvert^{-1}
\end{align}
where $\mathrm{det}$ is the determinant and $\bb{J}^{j}$ is the Jacobian of $g^{j}$ . We wish to learn $g^{j}$ such that the distribution of latent variables matches a standard Gaussian, i.e. $p(\bb{\hat{y}}^{j})=p(g^{j}(\bb{y}^{j}))\approx\mathcal{N}(0,\mathbf{I})$. To ensure $g^{j}$ is invertible, we model them as invertible neural network models described in \cite{Real_nvp,Freia}, and also used to morph dynamical systems in \cite{Fast_diff_int}, with trainable parameters. We can then train $g^{j}$ from each view, by minimising the negative log-likelihood of $g^{j}(\bb{y}^{j})$ being drawn from a standard Gaussian over dataset $\mathcal{D}$. We arrive at the following loss \cite{flows_review}:
\begin{align}
\mathcal{L}=-\sum_{\bb{y}^{j}\in\mathcal{D}}\big\{\log p(g^{j}(\bb{y}^{j}))+\log\lvert\mathrm{det}\bb{J}(\bb{y}^{j})\lvert\big\}.
\end{align}
As the number of demonstrations is typically small, to prevent the densities from collapsing into a delta function during training, we inject a small Gaussian noise into the data. After training up normalizing flow density estimators for each view, we can obtain densities $p(x^{j},y^{j}|t)$ over each view by evaluating \cref{eqn:flow}. Next, we seek to find the regions in Cartesian space which correspond to pixel coordinates that have high density.

\subsection{Trajectory Distribution Fitting via Ray-Tracing}\label{subsec:trajectory_fitting}
\begin{figure}
  \centering
  \begin{subfigure}[b]{0.23\textwidth}
    \fbox{\includegraphics[width=\textwidth]{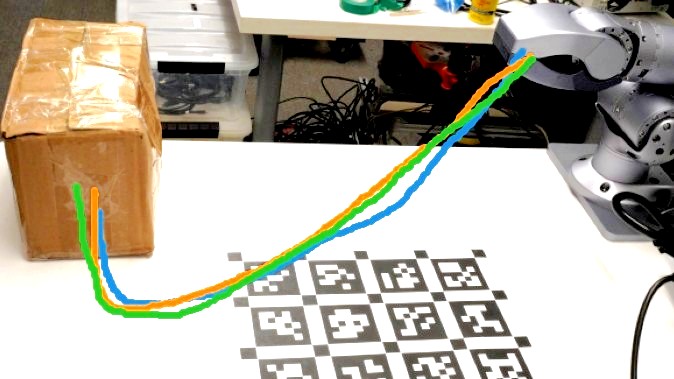}}
  \end{subfigure}%
  \begin{subfigure}[b]{0.23\textwidth}
    \fbox{\includegraphics[width=\textwidth]{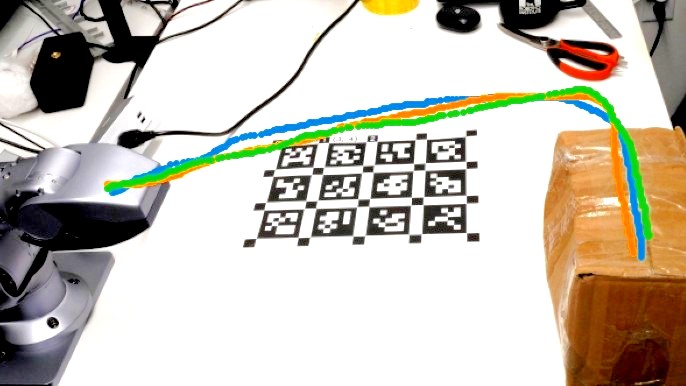}}
  \end{subfigure}%

      \begin{subfigure}[b]{0.153\textwidth}
    \fbox{\includegraphics[width=\textwidth]{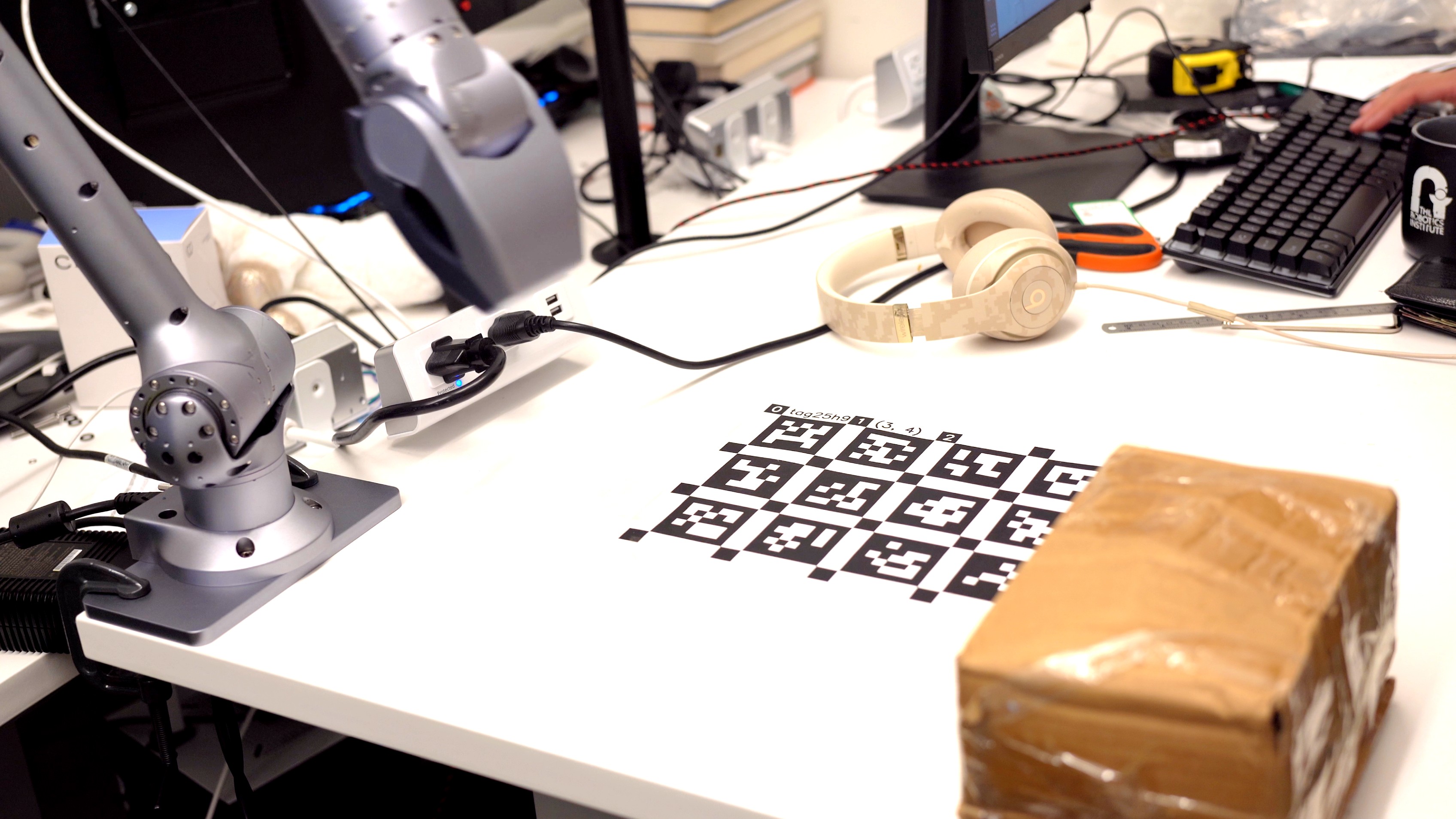}}
  \end{subfigure}%
    \begin{subfigure}[b]{0.153\textwidth}
    \fbox{\includegraphics[width=\textwidth]{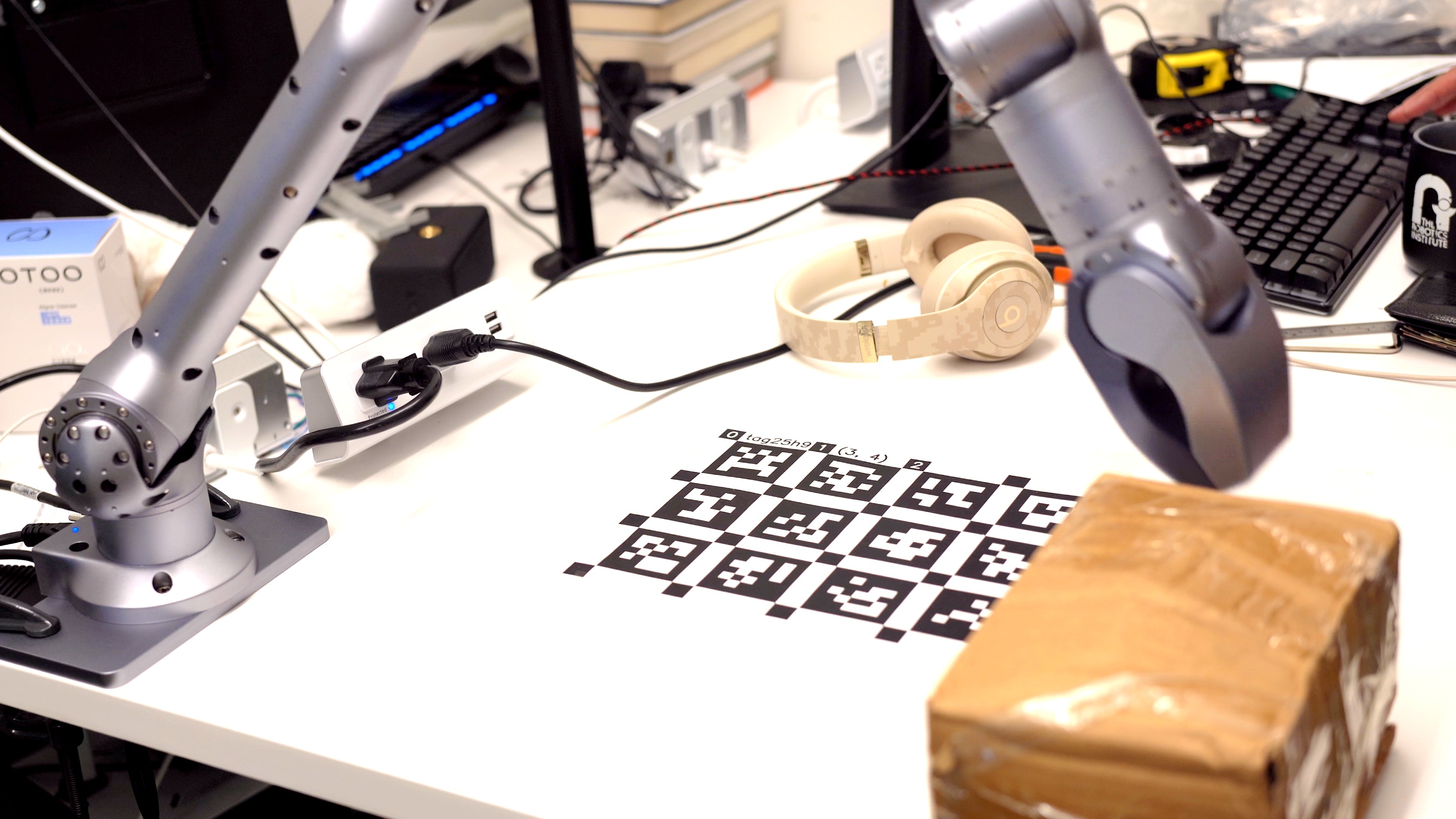}}
  \end{subfigure}%
      \begin{subfigure}[b]{0.153\textwidth}
    \fbox{\includegraphics[width=\textwidth]{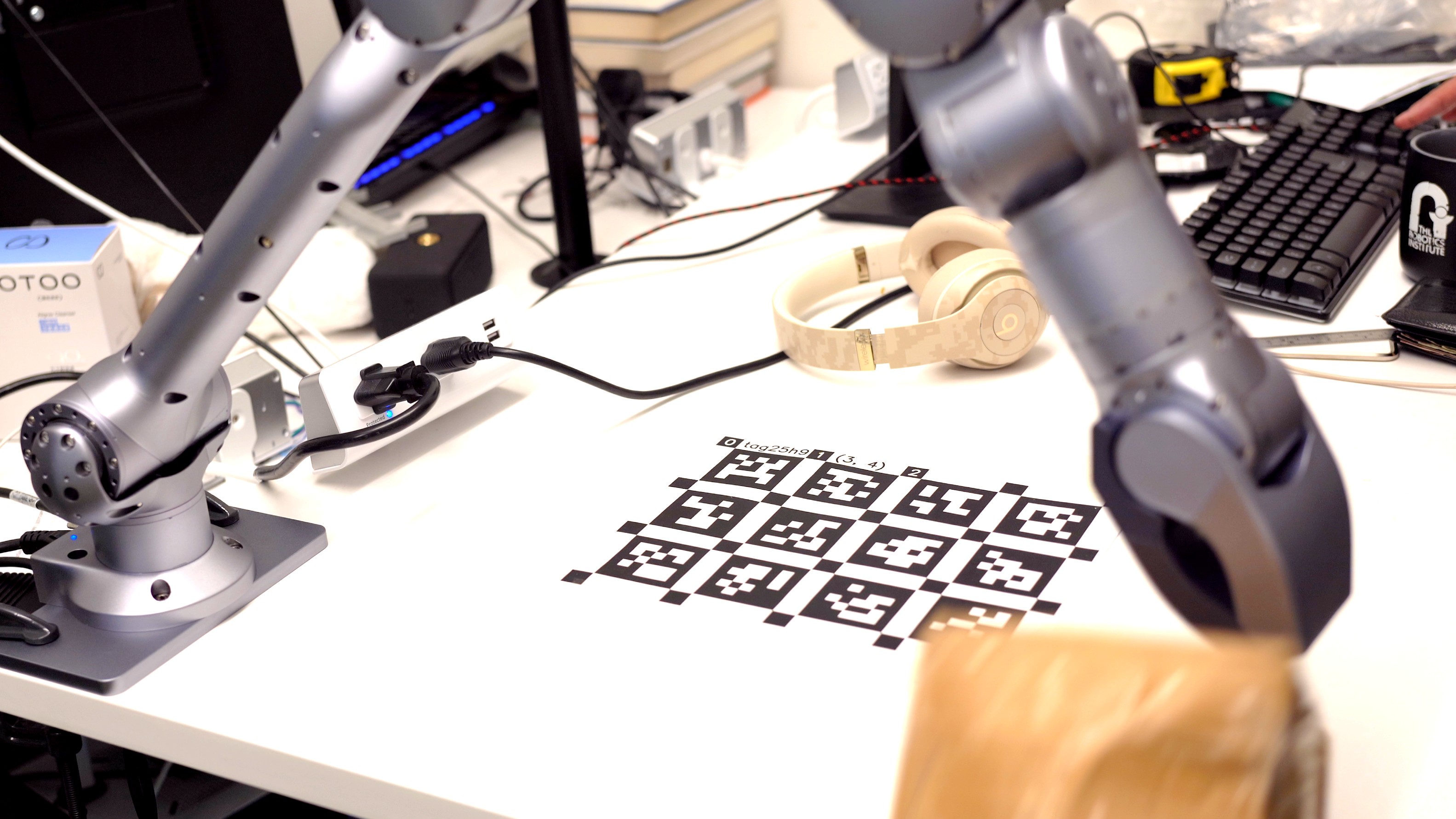}}
  \end{subfigure}%
  
  \begin{subfigure}[b]{0.46\textwidth}
    \fbox{\includegraphics[width=0.333\textwidth]{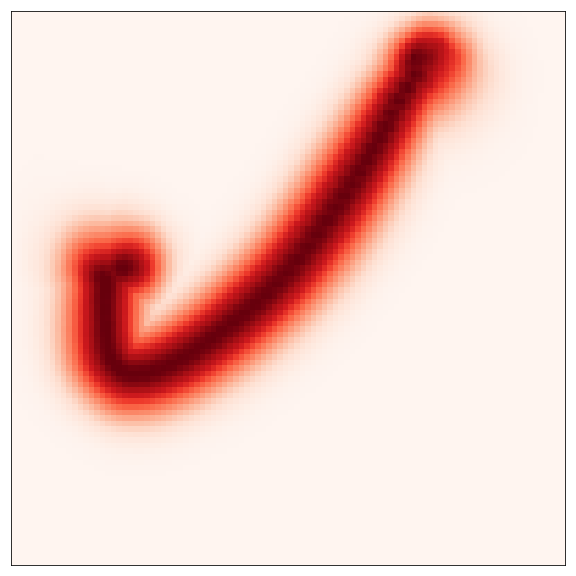}%
    \includegraphics[width=0.333\textwidth]{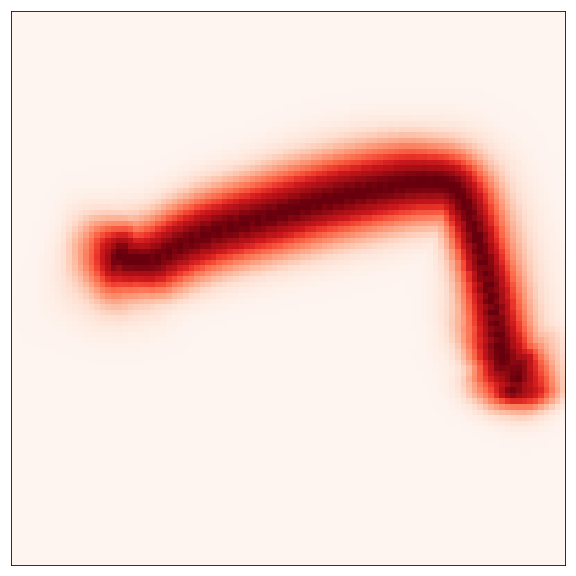}%
    \includegraphics[width=0.333\textwidth]{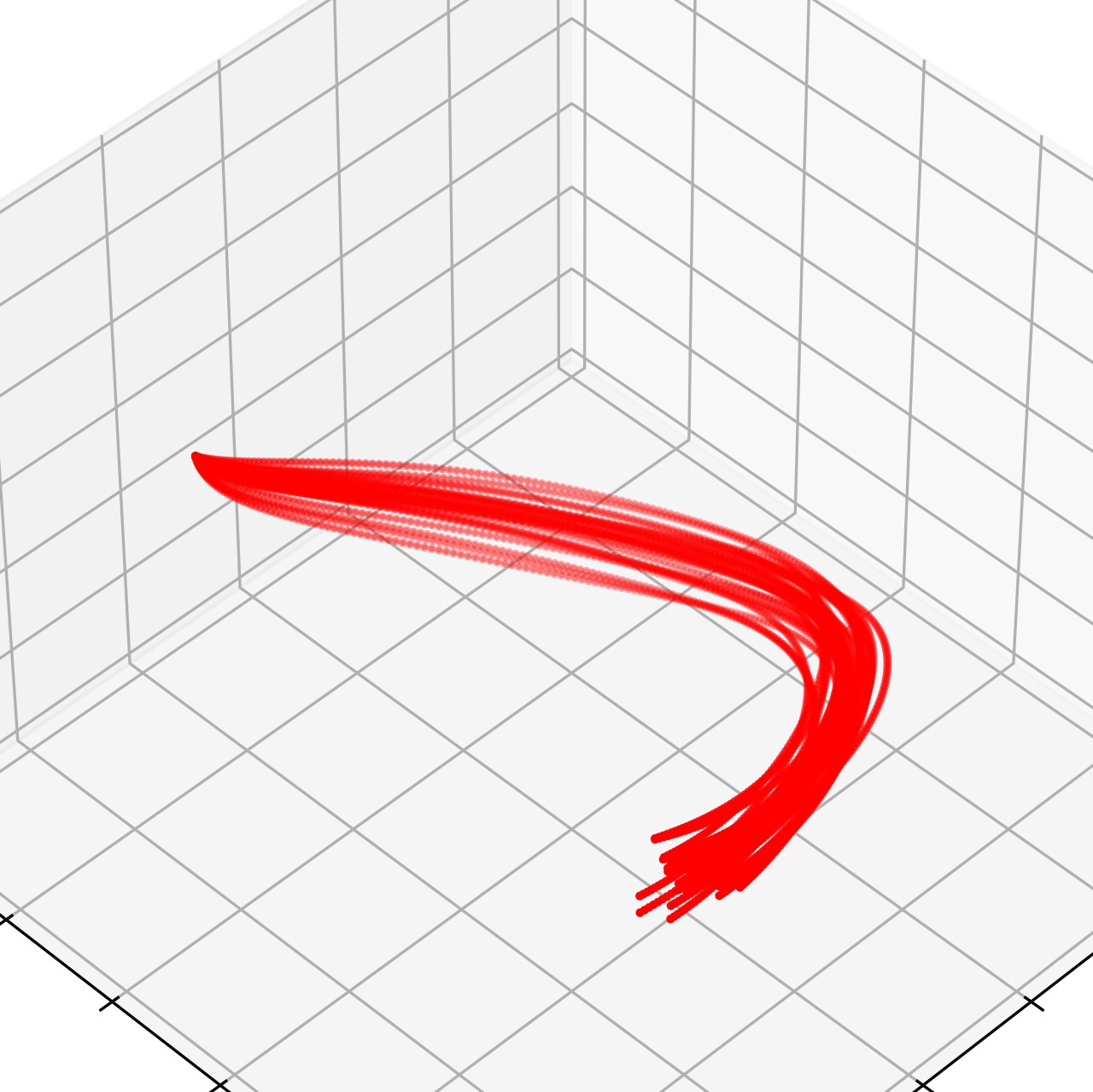}}
  \end{subfigure}%
  \caption{Example of learning to push a box. \underline{Top}: sketched demonstrations over two different views. \underline{Middle}: Robot executing the learned skill. \underline{Bottom}: Densities in 2D view space with time axis collapsed and trajectories from the model in 3D task space.}
  \label{fig:2d_to_3d}
\end{figure}

We apply ray-tracing along the 2D pixels of high estimated density from both views to find regions in 3D space where the rays intersect. We can then fit our generative model onto these regions. In this section, we shall (1) introduce the parameterisation of our generative model, and then (2) elaborate on finding and fitting the model on corresponding regions.
An example of learning to push a box is provided as \cref{fig:2d_to_3d}, where the view space density and the resulting distribution over trajectories are visualised.

\textbf{Distributions of Trajectories:} Similar to \cite{KTM, OTNet, Prob_struct_const}, generated motion trajectories are represented as functions $\bm{\xi}:t\rightarrow\bb{x}$, where $t\in[0,1]$ is a normalised time variable and $\bb{x}\in\mathbb{R}^{3}$ is a corresponding location. Trajectories are modelled by a linear combination of $M$ basis functions $\phi_{i}:t\rightarrow\mathbb{R}$ for $i=1,\ldots,M$, and are parameterised by a matrix of weights $\bb{W}\in\mathbb{R}^{M\times 3}$. We have:
\begin{align}
    \bm{\xi}(t)&=\bb{W}^{\top}\Phi(t), && \Phi(t)=[\phi_{1}(t),\ldots,\phi_{M}(t)]^{\top}\label{eqn:forward_traj}
\end{align}
where $\Phi$ is a $M$-dimension feature vector with each basis function evaluated once for each spatial dimension in Cartesian space. The basis functions can be selected to enforce \emph{a priori} assumptions on the motion trajectories, such as smoothness of periodicity. In this work, each basis is a squared exponential function that enforces the smoothness of motion, specifically, 
\begin{align}
\phi_{i}(t)=\exp{(-\gamma\lvert\lvert t-t_{i}\lvert\lvert_{2}^{2}}),
\end{align}
for $i=1,\ldots, M$, where $\gamma$ is a length-scale hyper-parameter where smaller values correspond to smoother trajectories. The $M$ times, $t_{i}$, are evenly spaced values between $0$ and $1$.

We can extend our parameterisation of a single trajectory to a \emph{distribution} of trajectories $p(\bm{\xi})$, by estimating a distribution over the weight matrix $\bb{W}$. For tractability, we assume independent Gaussian distributions over each element in $\bb{W}$. Let us denote each element in $\bb{W}$ as $w_{m,n}$ where $m=1,\ldots,M$ and $n=1,2,3$, where $n$ is each spatial dimension. Then, the joint distribution over $\bb{W}$ is simply the product of the distribution over each element,
\begin{align}
p(\bb{W})=\prod_{m=1}^{M}\prod_{n=1}^{3}p(w_{m,n})=\prod_{m=1}^{m}\prod_{n=1}^{3}\mathcal{N}(\mu_{m,n},\sigma_{m,n}^{2}), \label{eqn:gaussian_weights}
\end{align}
where the means and standard deviations of the distribution over each $w_{m,n}$ are denoted as $\mu_{m,n}$ and $\sigma_{m,n}$. Fitting the distribution of trajectories involves finding each $\mu_{m,n}$ and $\sigma_{m,n}$ to match the given data.

\begin{algorithm}[t]
\scriptsize
\caption{{Sample from Intersecting Regions}}\label{alg:intersect}
\SetKwInOut{Input}{input}
\SetKwInOut{Output}{output}
\SetKwInOut{Initalise}{initalise}
\newcommand{\lIfElse}[3]{\lIf{#1}{#2 \textbf{else}~#3}}
\SetKwFor{DoParallel}{Do in Parallel for each (}{) $\lbrace$}{$\rbrace$}
\Input{Densities $p(x^{j},y^{j}|t)$, ray-tracing function $f_{r}$, grid points $G=\{(x^{j},y^{j})_{i}\}_{i=1}^{N_{G}}$, distances $D=\{d_{1},\ldots,d_{N_{d}}\}$, times $T=\{t_{1},\ldots,t_{N_{t}}\}$, density and distance thresholds $\epsilon, \delta$.}

$intersections\_over\_t \leftarrow \{\}$;\\
\For{$t \in T$}{
$R^{1}_{t}\cap R^{2}_{t}\leftarrow \{\}$, $R^{1}_{t}\leftarrow\{\}$, $R^{2}_{t}\leftarrow\{\}$;\\
\For{$j\in\{0,1\}$}{
\For{$(x^{j},y^{j})\in G$}{
\uIf{$p(x^{j},y^{j}|t)\geq \epsilon$}{
{\color{blue} Ray-trace for $x^{j},y^{j}$ above threshold:}\\
\For{$d\in D$}
{
$\bb{x}\leftarrow f_{r}(d,x^{j},y^{j})$;\\
$R^{j}_{t}.\mathrm{insert}(\bb{x})$;\\
}
}
\textbf{end}}
}
{\color{blue} Intersecting if pairwise distances below $\delta$}\\
$R^{1}_{t}\!\cap\! R^{2}_{t}.\mathrm{insert}(\bb{x}_{1}\!,\!\bb{x}_{2}), 
\forall \bb{x}_{1},\bb{x}_{2}\in R^{1}_{t} \!\times\! R^{2}_{t}, \text{ such that } \lvert\lvert \!\bb{x}_{1}\!-\!\bb{x}_{2} \!\lvert\lvert_{2}\!<\!\delta$;\\
$intersections\_over\_t.\mathrm{insert}(R^{1}_{t}\cap R^{2}_{t})$;
}
\Output{$intersections\_over\_t$}
\end{algorithm}

\begin{wrapfigure}{L}{0.16\textwidth}
\centering
\fbox{\includegraphics[width=0.16\textwidth]{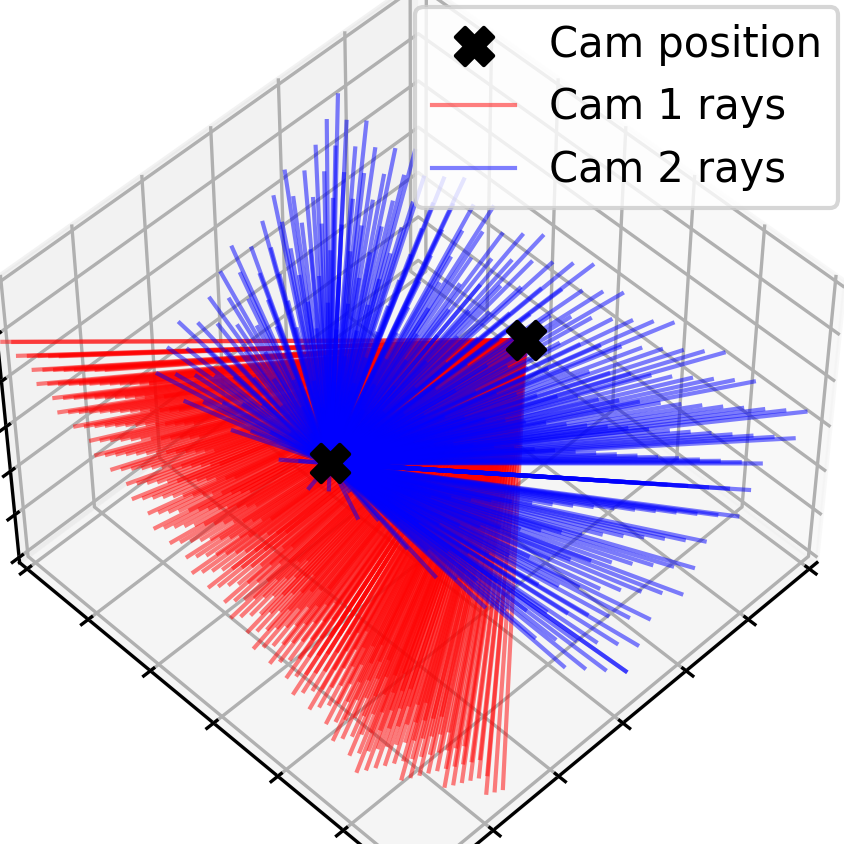}}
\caption{Rays traced from cameras at different poses. }\label{fig:example_ray}
\end{wrapfigure}

\textbf{Ray-tracing from view space densities:} Provided time-varying densities over pixels from different views, $p(x^{j},y^{j}|t)$ for $j\in\{1,2\}$, we wish to fit a trajectory distribution $p(\bm{\xi})$ by tracing the path of rays. We follow classical rendering methods \cite{ray_tracing_course} used for NeRF models \cite{mildenhall2020nerf} and assume pin-hole cameras at each view. We construct the ray in 3D which passes through each coordinate in 2D view space as, 
\begin{equation}
f_{r}(d,x^{j},y^{j})=\bb{o}^{j}+\bm{\omega}(x^{j},y^{j})d,
\end{equation}
where $\bb{o}^{j}$ is the origin of the camera, $\bm{\omega}(x^{j},y^{j})$ is a direction, and $d$ is a distance bounded between $d_{near}$ and $d_{far}$. These values are directly obtainable from the camera specifications. \Cref{fig:example_ray} shows an example of rays traced from two cameras at different poses. From each view, the region corresponding to the density above threshold $\epsilon$ at a given time $t$ is the codomain of $f_{r}$. Let us define this as the set: 
\begin{align}
R^{j}_{t}=\{\bb{x}\in\mathbb{R}^{3}|&f_{r}(d,x^{j},y^{j}),\nonumber\\
&\text{for all }d\in[d_{near},d_{far}]\text{ and }x^{j}, y^{j}\in[0,1],\nonumber\\
&\text{such that }p(x^{j},y^{j}|t)\geq \epsilon\}.
\end{align}

We seek the intersection of the 3D regions which corresponds to densities from both views, i.e. $R^{1}_{t}\cap R^{2}_{t}$. This can be approximated by sampling regular-spaced grid points over $d$ and view space coordinates $(x^{j},y^{j})$, for both views. If the distance between a sample from one view and its closest sample from the other view is below a specified distance threshold $\delta$, then we consider both samples to be in the intersecting region. An outline of drawing samples from the intersecting region is presented in \cref{alg:intersect}. In practice, the nested loops are vectorised and can be efficiently executed on GPUs using deep learning frameworks, such as PyTorch \cite{pytorch}. We can obtain a set of $n_{s}$ intersecting 3D spatial coordinates along the normalised time $t$, i.e. $\mathcal{S}=\{(t_{i},\bb{x}_{i})\}_{i=1}^{n_{s}}$.

To fit our generative model on $\mathcal{S}$, we need to compute the time-conditional distributions in Cartesian space, i.e. $p(\bm{\xi}|t)$. Let us first stack our train-able mean and variance parameters from \cref{eqn:gaussian_weights}, 
\begin{align}
\mathcal{M}=
\begingroup 
\setlength\arraycolsep{2pt}
\begin{bmatrix}
    \mu_{1,1}&\mu_{1,2}&\mu_{1,3}\\
    \vdots & \vdots & \vdots\\
    \mu_{M,1}&\mu_{M,2}&\mu_{M,3}
\end{bmatrix}\endgroup &&
\bm{\Lambda}=
\begingroup 
\setlength\arraycolsep{2pt}
\begin{bmatrix}
    \sigma^{2}_{1,1}&\sigma^{2}_{1,2}&\sigma^{2}_{1,3}\\
    \vdots & \vdots & \vdots\\
    \sigma^{2}_{M,1}&\sigma^{2}_{M,2}&\sigma^{2}_{M,3}
\end{bmatrix}.\endgroup \nonumber
\end{align}
As the weight distribution $p(\bb{W})$ is Gaussian and involves a linear transformation with $\Phi$, as given in \cref{eqn:forward_traj}, we have:
\begin{align}
p(\bm{\xi}|t)=\mathcal{N}(\mathcal{M}^{\top}\Phi(t),\mathrm{Diag}(\bm{\Lambda}^{\top}\Phi(t)^{2})),
\end{align}
where $\mathrm{Diag}(\cdot)$ produces a square matrix with the inputted vector as its diagonal. We minimise the Gaussian negative log-likelihood of $p(\bm{\xi}|t)$ over points in $\mathcal{S}$ to fit the parameters $\mathcal{M}$ and $\bm{\Lambda}$.

\begin{figure}[t]
  \centering 
  \begin{subfigure}[b]{0.297\textwidth}
    \fbox{\includegraphics[width=0.333\textwidth]{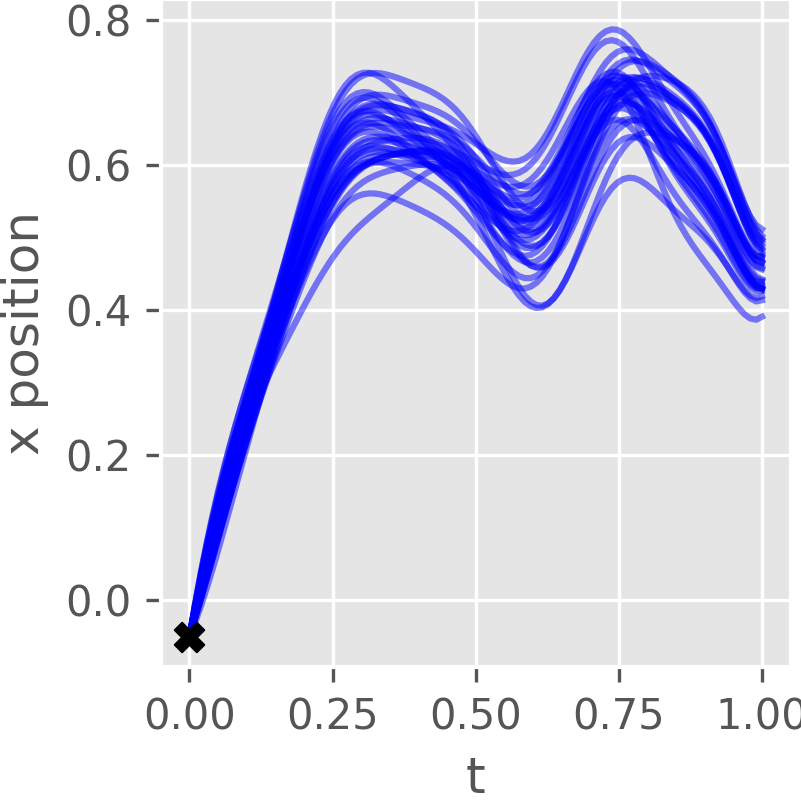}%
    \includegraphics[width=0.333\textwidth]{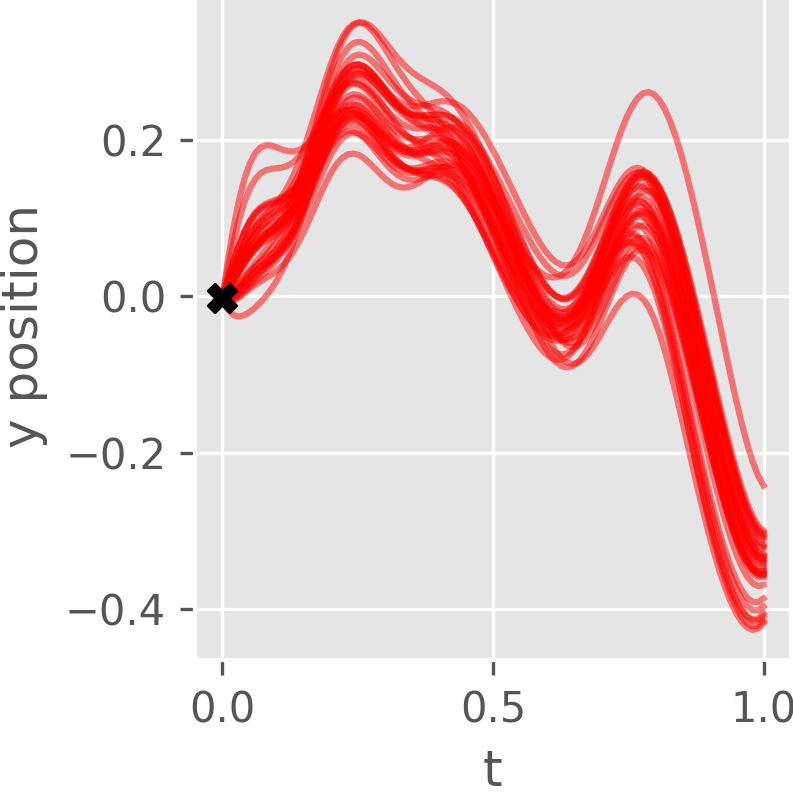}%
    \includegraphics[width=0.333\textwidth]{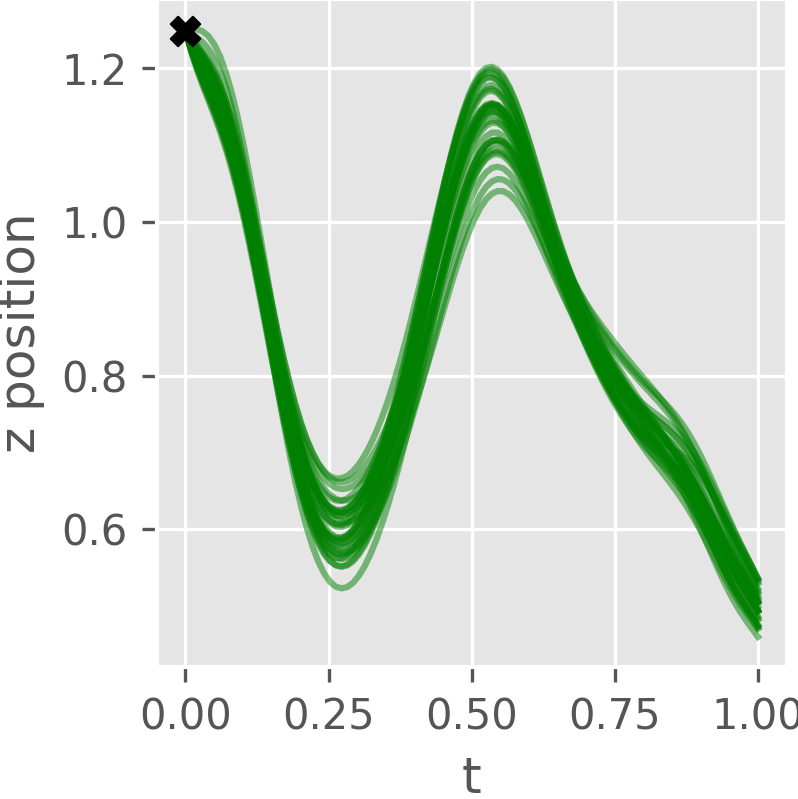}}%
  \end{subfigure}%
\begin{subfigure}[b]{0.199\textwidth}
    \fbox{\includegraphics[width=\textwidth]{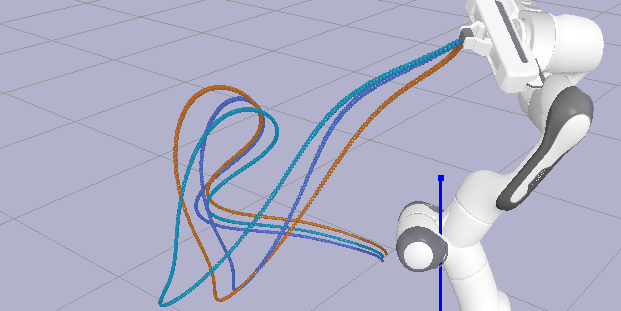}}
  \end{subfigure}
  \caption{We use Diagrammatic Teaching to teach the robot to follow ``R'' characters. \underline{Left}: The $x,y,z$ positions of sampled trajectories over normalised time $t$. The initial positions of the trajectory samples are enforced at the black marker. \underline{Right}: Three end-effector trajectories, conditioned to start from the current position.}
  \label{fig:positions}
\end{figure}

\subsection{Conditional Trajectory Generation at New Positions}\label{subsec:trajector_gen}
Trajectories learned by RPTL can be further adapted to start and new starting positions. After training to obtain fitted means, $\mathcal{M}$, and variances, $\bm{\Lambda}$, we can generate a collection of trajectories by sampling elements in $\bb{W}$ from $w_{i,j}\sim\mathcal{N}(\mu_{i,j},\sigma_{i,j}^{2})$, and then evaluate \cref{eqn:forward_traj}. However, often times we have additional knowledge of where the trajectory shall be at the starting position. For example, the generated task space trajectory at the initial time should match its current end-effector position ($\bb{x}_{eef}$), i.e. enforcing, $\bm{\xi}(0)=\bb{x}_{eef}$.

Let us begin by defining notation: let $\bb{W}_{1}$ be the first row of $\bb{W}$, and $\bb{W}_{2:}$ be the others; let $\Phi(0)_{1}$ be the first element in $\Phi(0)$ and $\Phi(0)_{2:}$ denote the others. For a specific sampled trajectory, we draw a sample of $\bb{W}$ and then alter $\bb{W}_{1}$ from the enforced condition. Specifically, at the beginning of the trajectory, we wish to enforce:
\begin{align}
    \bm{\xi(0)}=\bb{W}_{1}^{\top}\Phi(0)_{1}+\bb{W}_{2:}^{\top}\Phi(0)_{2:}=\bb{x}_{eef},
\end{align}
allowing us to solve for $\bb{W}_{1}$, together with the known $\bb{W}_{2:}$. An example of generated trajectories conforming to starting at the current end-effector position is shown in \cref{fig:positions}, where the robot is taught to sketch out ``R'' characters, along with the $x,y,z$ positions of generated trajectories over $t$.

\begin{figure*}[t]
  \centering 
  \begin{subfigure}[b]{0.33\textwidth}
    \fbox{\includegraphics[width=0.499\textwidth]{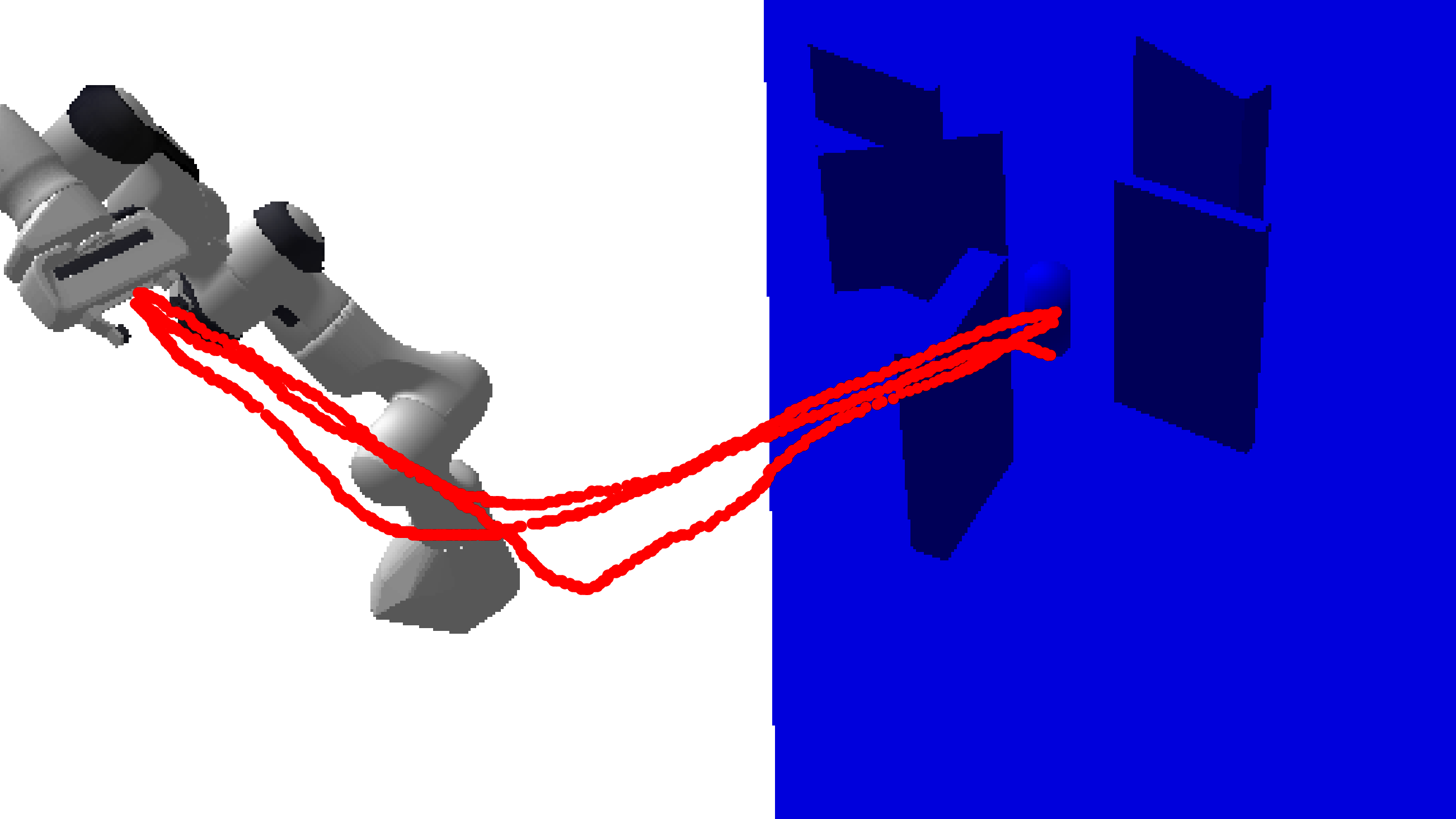}%
\includegraphics[width=0.499\textwidth]{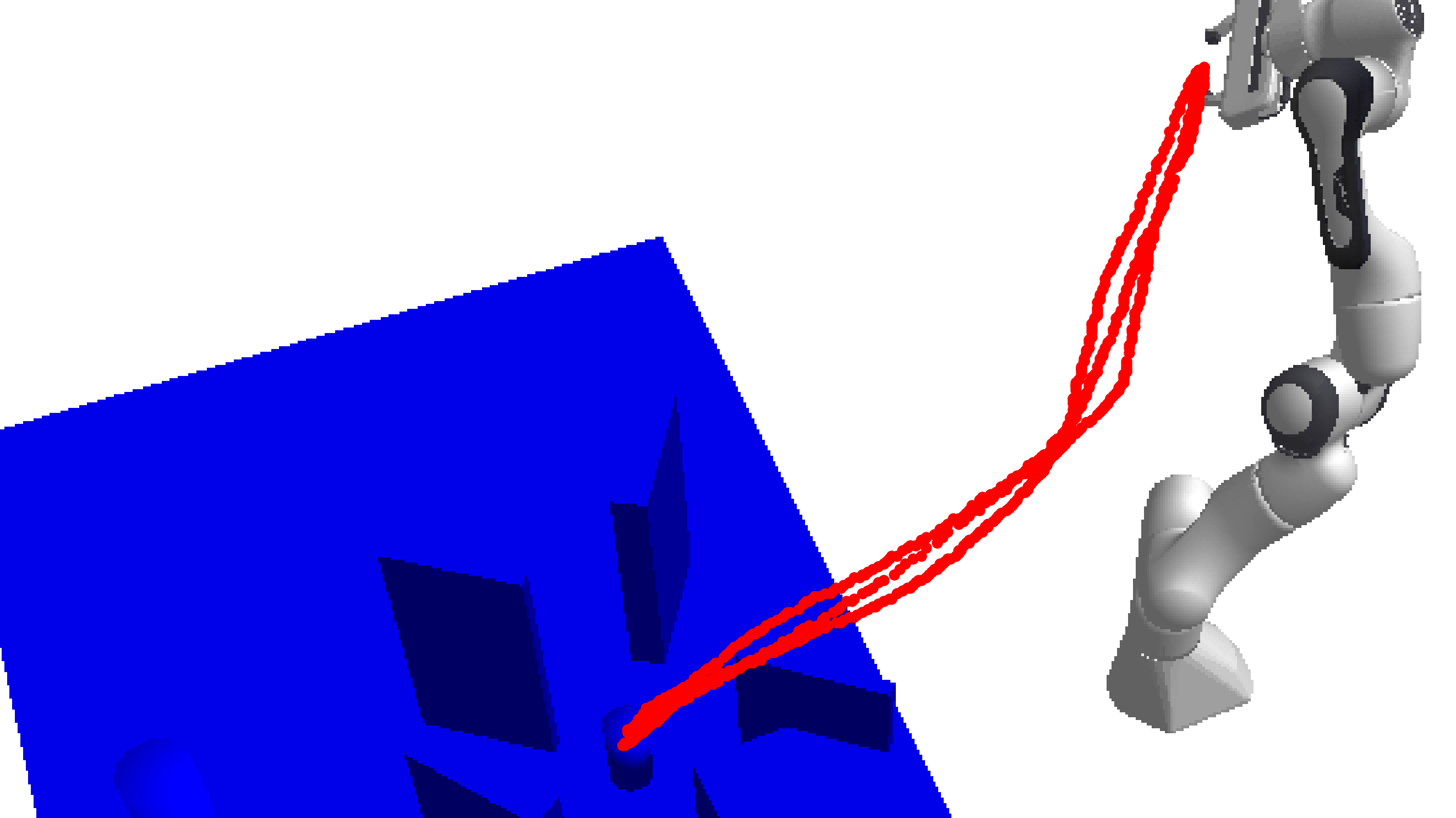}}%
    
    \fbox{\includegraphics[width=\textwidth]{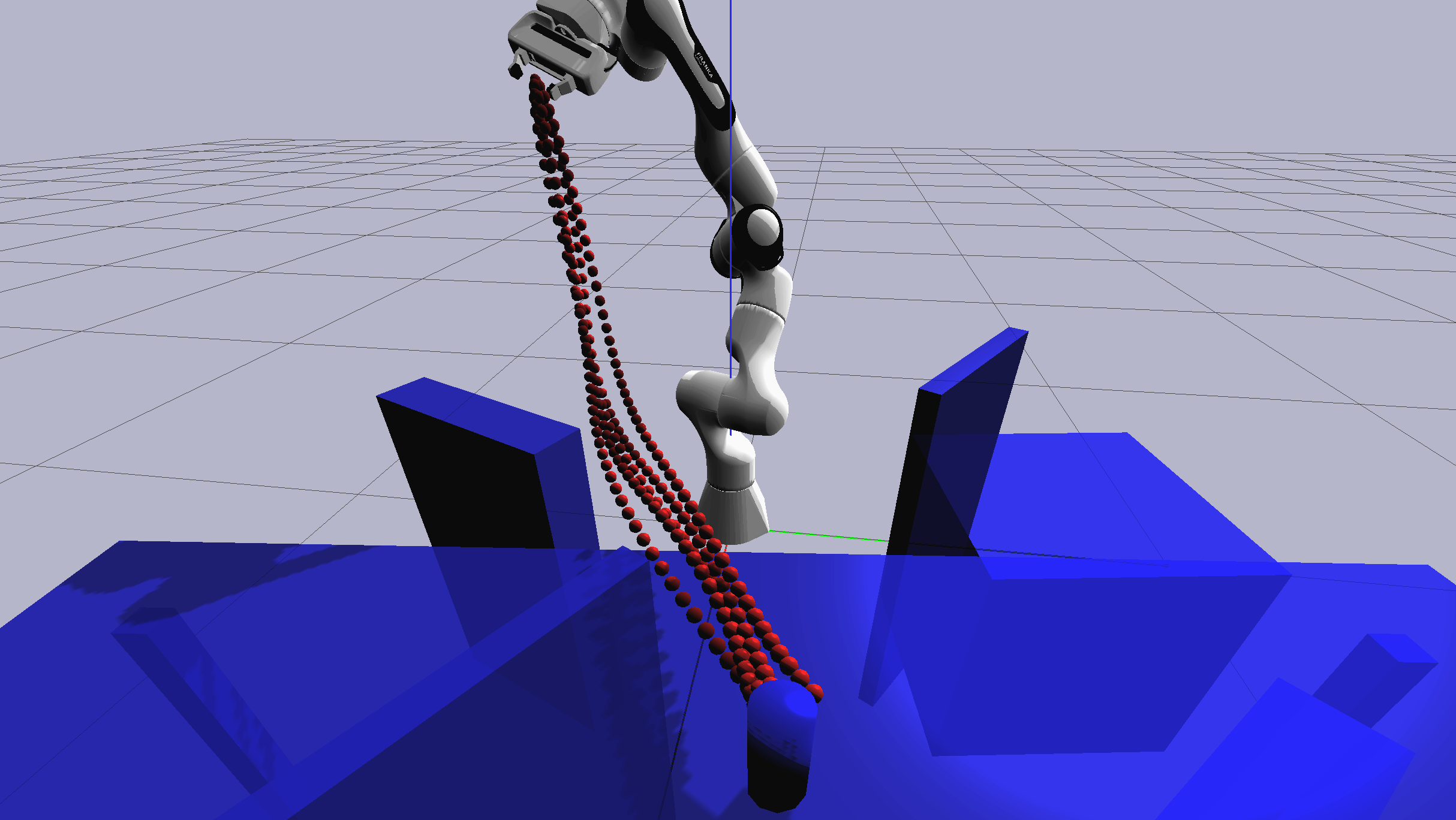}}%
  \end{subfigure}%
  \begin{subfigure}[b]{0.33\textwidth}
    \fbox{\includegraphics[width=0.499\textwidth]{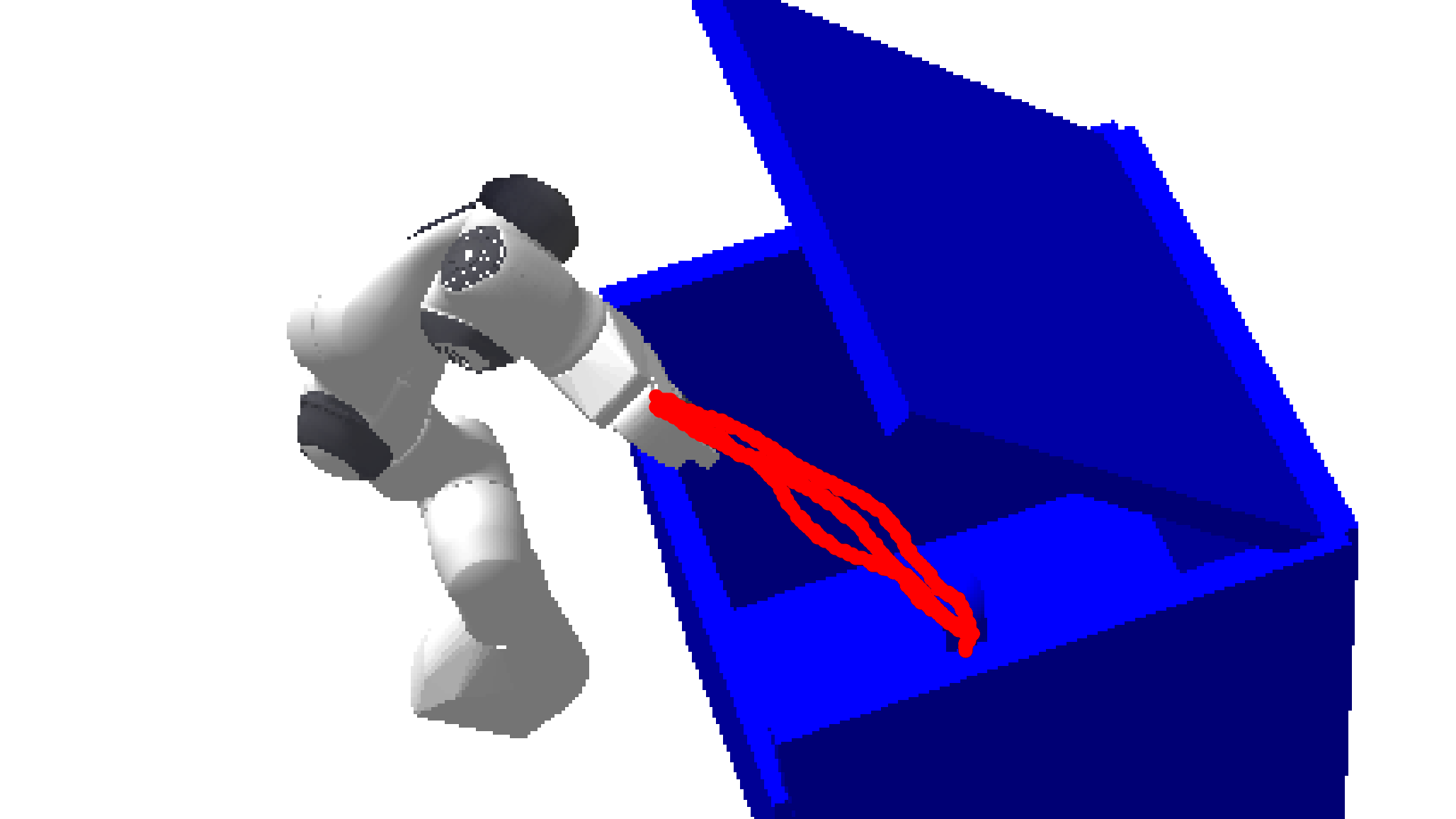}%
\includegraphics[width=0.499\textwidth]{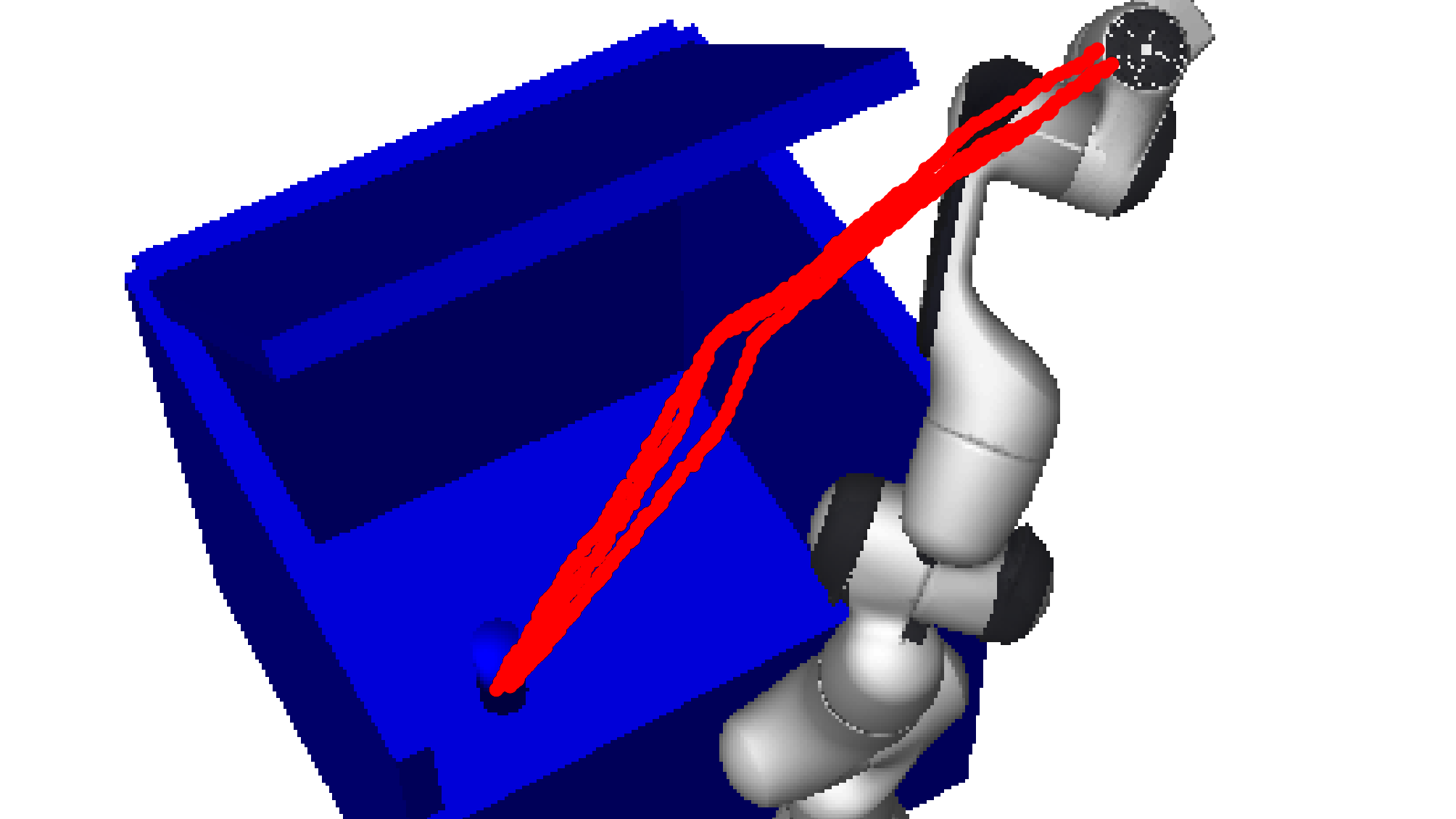}}%
    
    \fbox{\includegraphics[width=\textwidth]{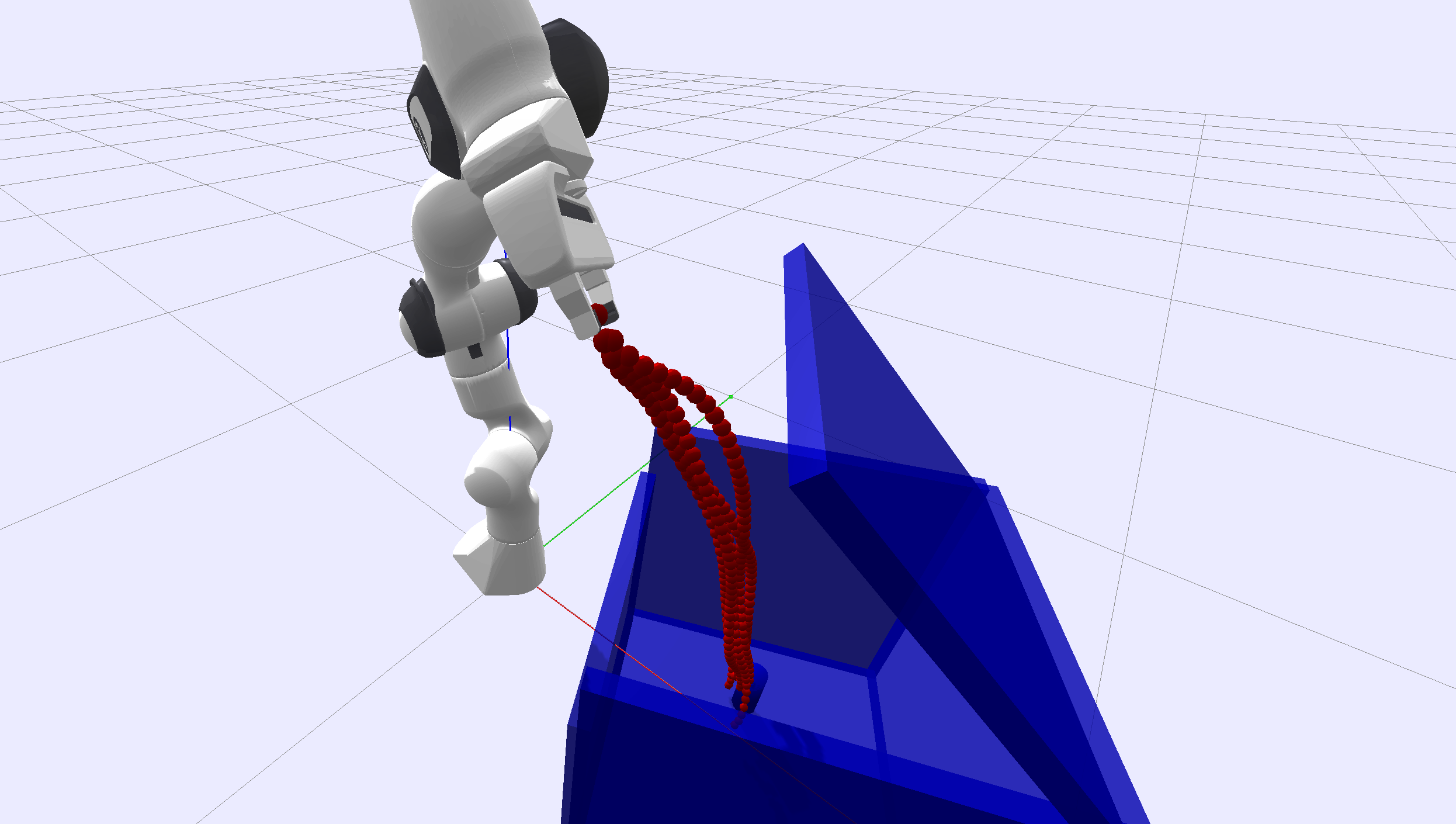}}%
  \end{subfigure}%
    \begin{subfigure}[b]{0.33\textwidth}
    \fbox{\includegraphics[width=0.499\textwidth]{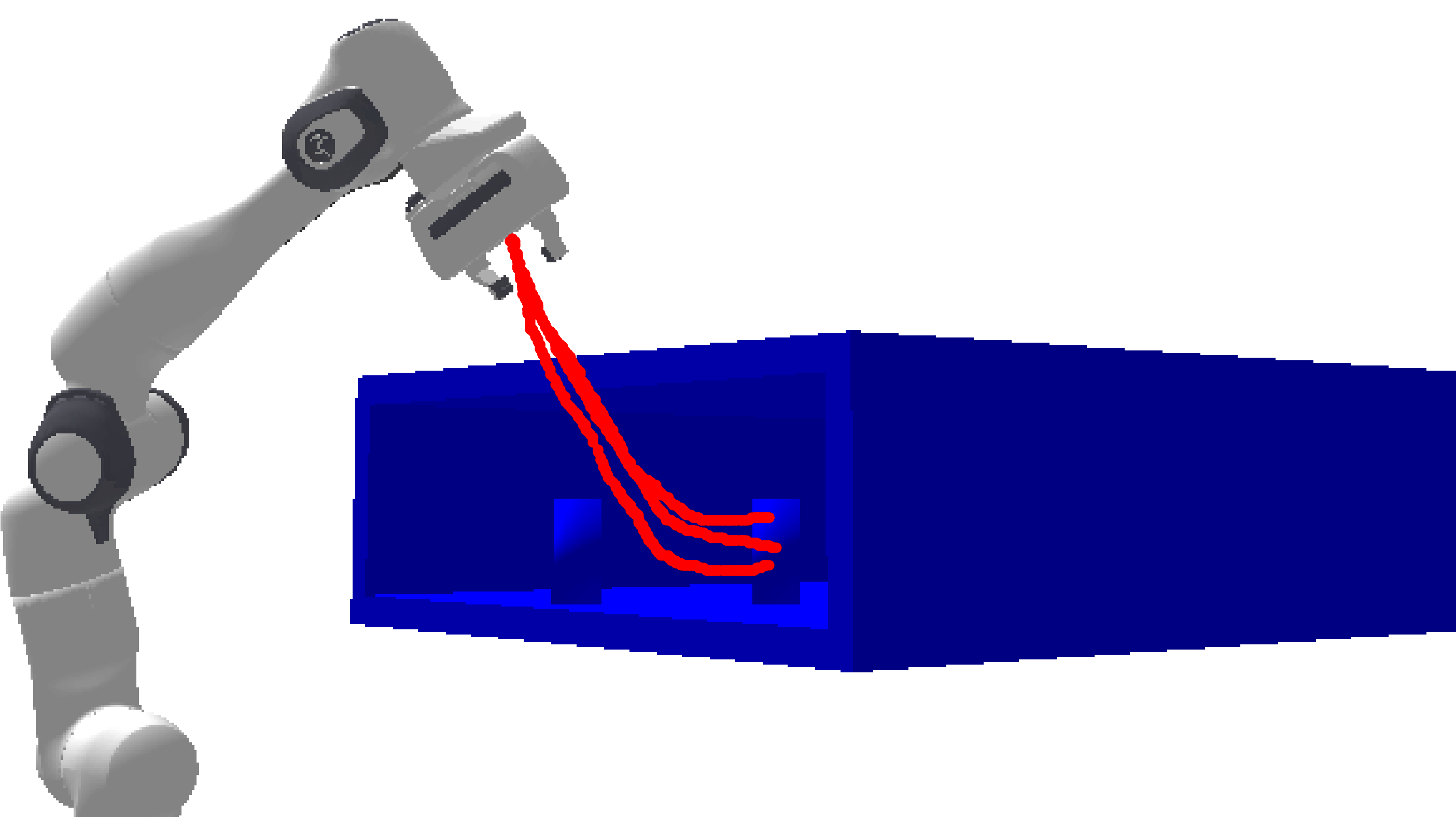}%
\includegraphics[width=0.499\textwidth]{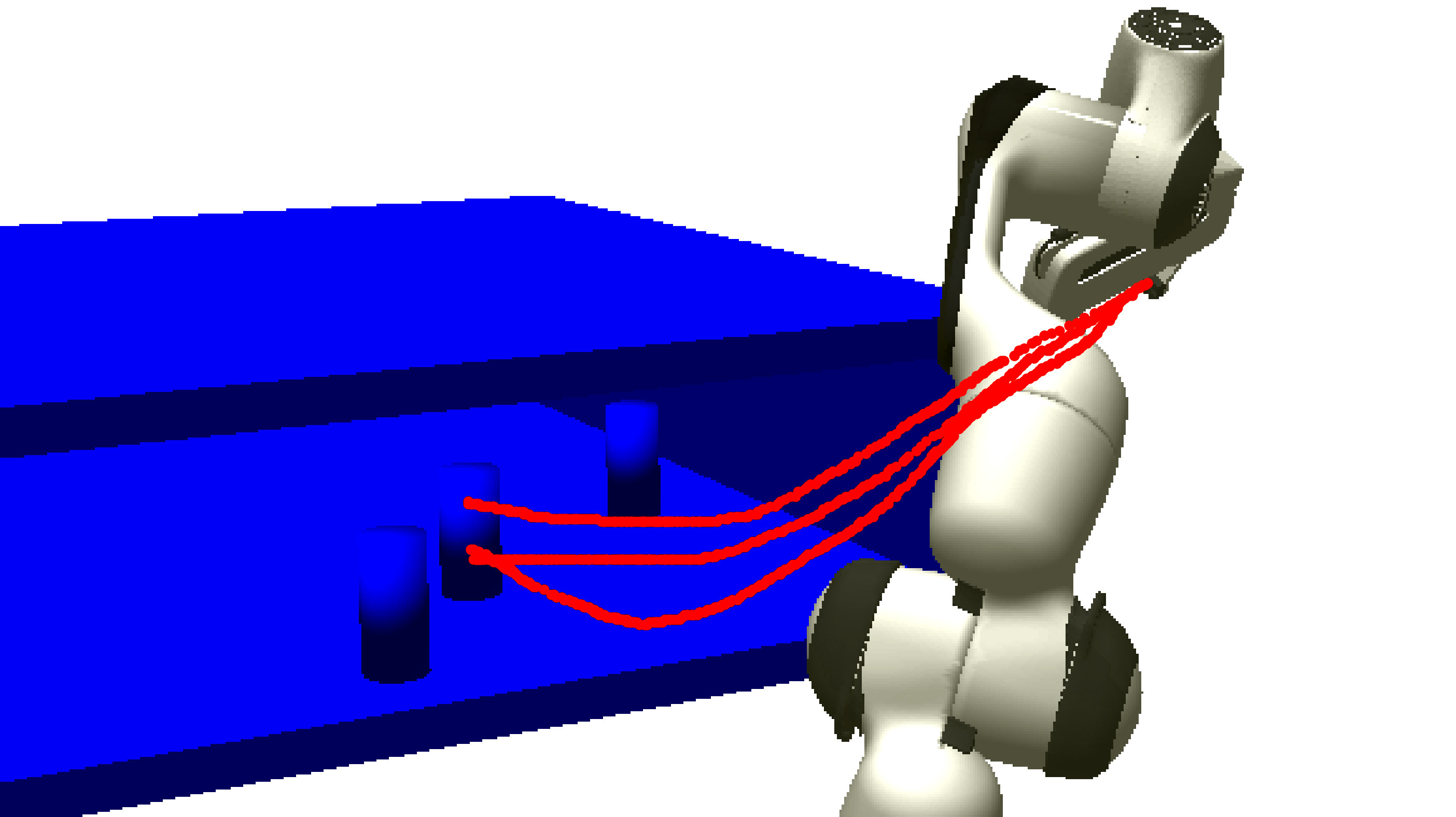}}%
    
    \fbox{\includegraphics[width=\textwidth]{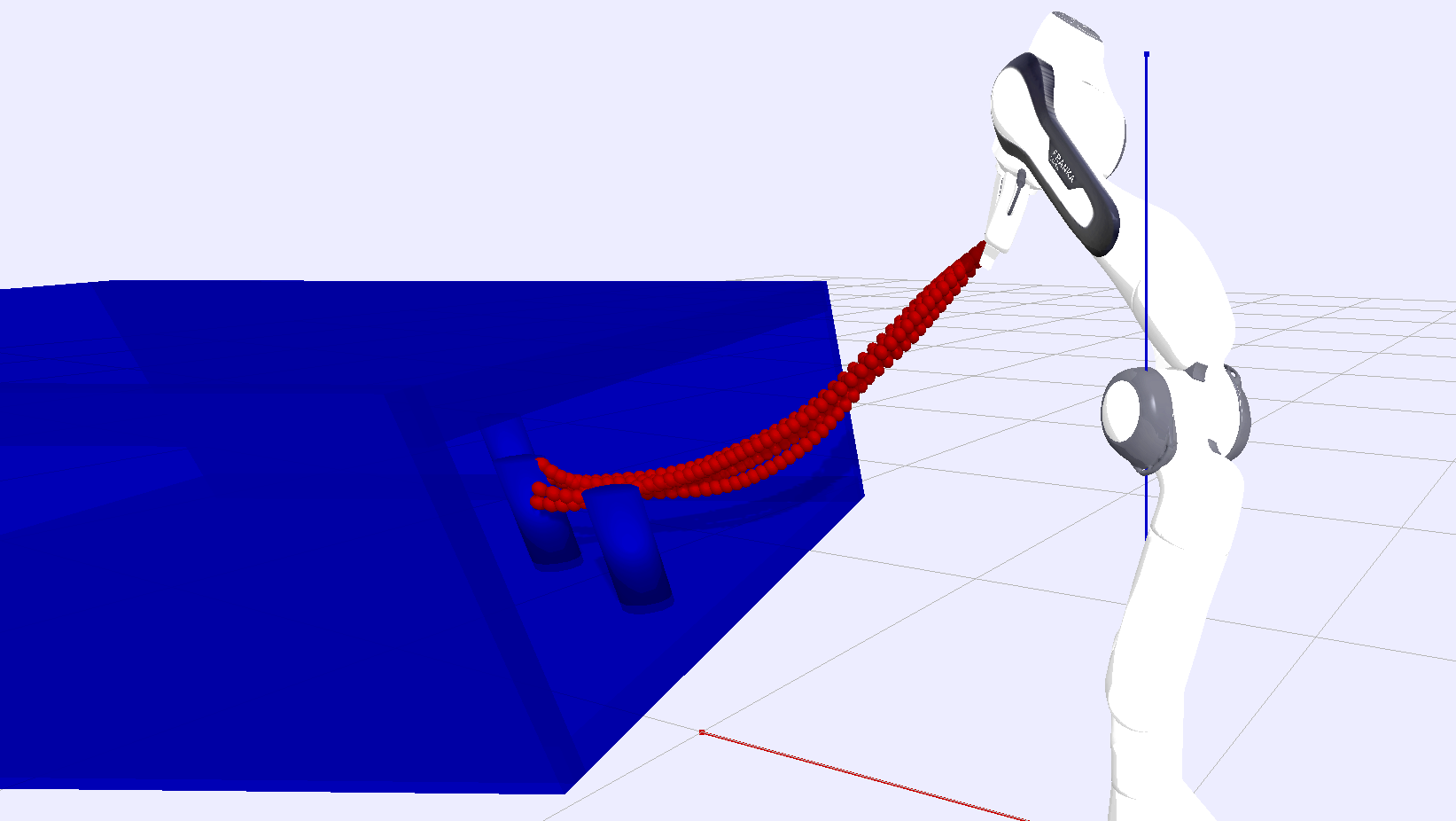}}%
  \end{subfigure}%
  \caption{We teach the manipulator motion in the \emph{table top}, \emph{box}, and \emph{shelf} environments (\underline{left, middle, right} respectively). The \underline{top} row images are two camera views provided to the user, and the red trajectories are 2D demonstrations sketched by the user. We sample five 3D motion trajectories from the trajectory distribution model and illustrate them in red in the images on the \underline{bottom} row. We see that the learned distribution of trajectories in 3D Cartesian space is consistent with the user's sketches.}
  \label{fig:experiment_example}
\end{figure*}
\begin{figure*}[h]
  \centering 
    \fbox{\includegraphics[width=0.32\textwidth]{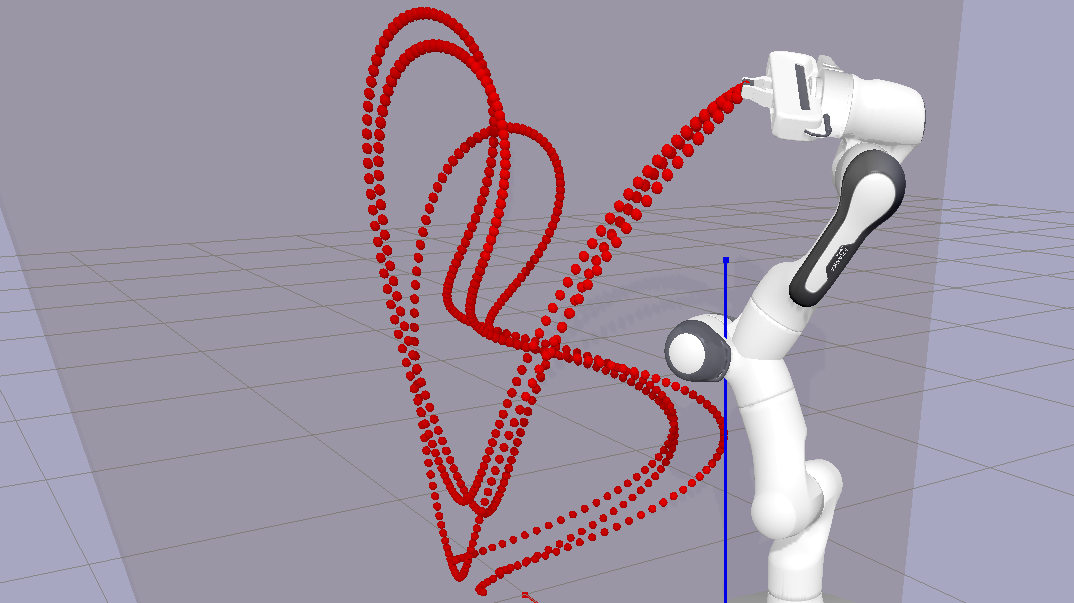}}%
    \fbox{\includegraphics[width=0.32\textwidth]{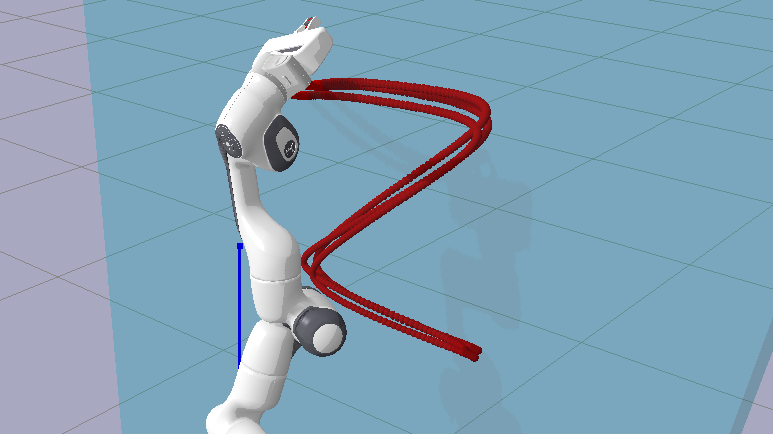}}%
    \fbox{\includegraphics[width=0.32\textwidth]{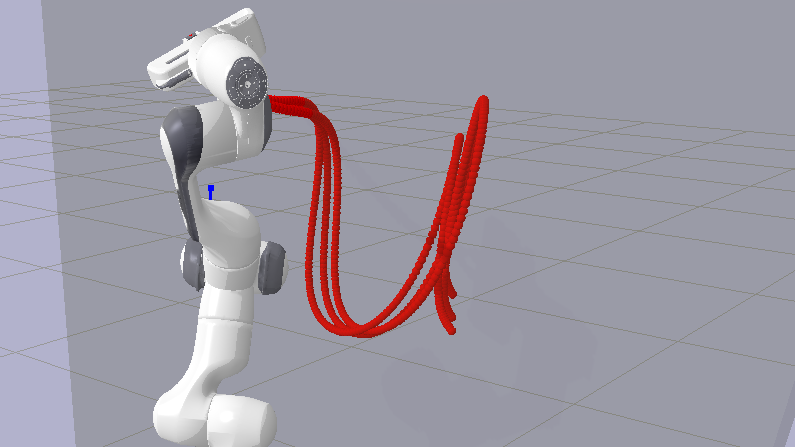}}%
  \caption{We diagrammatically teach the manipulator to sketch out the letters ``B'', ''Z'', ``U'', and sample three trajectories from the trained model. The model accurately generates the desired letters from randomised starting configurations. Example for ``R'' in \cref{fig:positions}.}
  \label{fig:letter_draw}
\end{figure*}
\section{Experimental Evaluation}\label{sec:experiments}
We empirically explore the performance of the proposed RPTL for the Diagrammatic Teaching problem in both simulation and real robot platforms. Specifically, we first examine the quality of the trained RPTL model and whether the motion acquired is consistent with user expectations. Then, we examine a case of using RPTL to diagrammatically teach the tracing of challenging alphabet characters, which takes several turns in movement and requires greater intricacy. Finally, we examine the robustness of RPTL on real-world robot platforms, by generating motion for a fixed-based 6 degrees-of-freedom manipulator and a quadruped robot with a mounted manipulator.

\subsection{Quality and Consistency of RPTL}
We set up three simulated environments with a Franka manipulator, following the \emph{table top}, \emph{box}, and \emph{shelf} environment types in \cite{Chamzas2022MotionBenchMakerAT}, using the physics simulator PyBullet \cite{coumans2019}. We place cameras at three different poses in the environment and ask the user to provide three demonstrations per camera view to reach a cylindrical goal object. The goal position is not given to the model, and the only specifications given are the diagrammatic demonstrations. We select two views and their corresponding demonstrations for training and hold out the demonstrations from the third view as a test set. To benchmark motion trajectories generated from the learned model, we project the generated 3D trajectories into the 2D view of the third view and compute distances between the projected trajectories and the retained test trajectories. Distances computed are all normalised by the width of the test image. By testing against a hidden test set, we can assess whether the produced motions match the expectation of the user and are consistent when viewed from a different pose.

\textbf{Metrics:} We use the following metrics to measure the quality of the learned trajectory distribution: (1)
 \emph{Mean Fr\'{e}chet distance (MFD)}: We compute the discrete Fr\'{e}chet distance \cite{eiter1994computing} between the mean of the trajectory distribution model and each of the test trajectory, then record the averages and standard deviations of the Fr\'{e}chet distances. The Fr\'{e}chet distance measures the distance between curves and can account for the ordering of points and handle varying lengths. (2) \emph{Wasserstein Distance (WD)}: We compute the 2-Wasserstein distance implemented in \cite{flamary2021pot} between five trajectories drawn from our model and the set of test trajectories. Crucially, the WD can measure distances between distributions beyond simply considering the mean of the probabilistic model.

 \textbf{Baselines:} We evaluate our model to the following baseline models. (1) \emph{Linear}: We provide the mean start and end positions of the test trajectories and assume a linear curve between them. Note that the other methods do not assume knowledge of the mean test start and end positions. (2) \emph{Nearest Neighbour (NN)}: We predict the trajectories as being identical to those traced from the nearest camera in the training set.
 
    
    
    
    

\begin{figure*}[!t]
  \includegraphics[width=\linewidth]{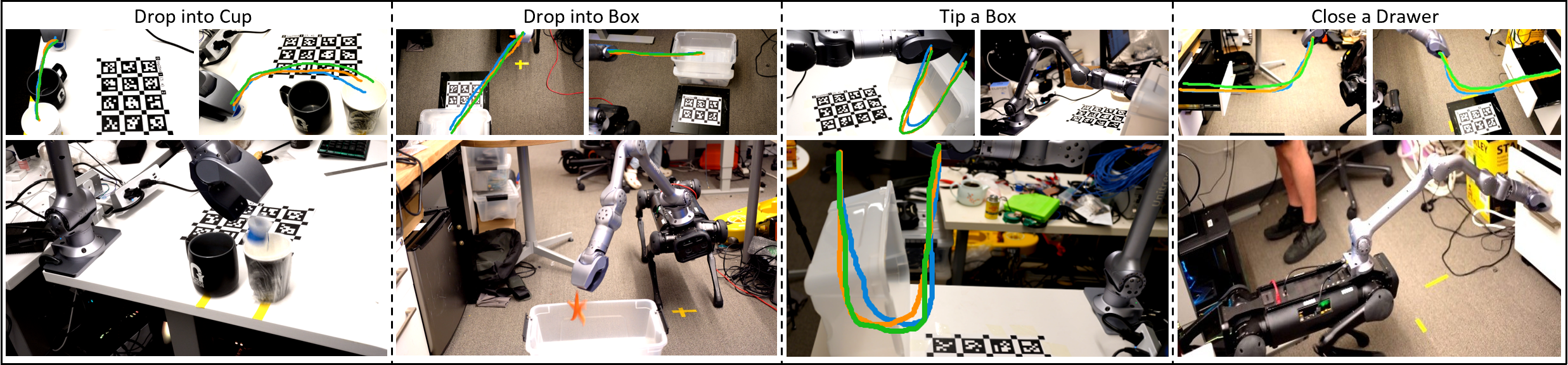}
  \caption{RPTL applied on real-world robot platforms. We diagrammatically teach the robot for the tasks \textbf{Drop into cup}, \textbf{Drop into box}, \textbf{Tip box}, and \textbf{Close a drawer}. The sketched demonstrations are shown in the \underline{top} row, while robot execution is shown at the \underline{bottom}. Additionally, the task \textbf{Push box} is shown in \cref{fig:2d_to_3d}.}
  \label{fig:real_world}
\end{figure*}

\setlength{\heavyrulewidth}{0.2em}
\begin{table}[t]
\centering
\caption{Comparison of RPTL and baselines on different environments. Lower distances indicate better performance.}
\label{tab:comparison}
\begin{adjustbox}{width=0.48\textwidth}
\begin{tabular}{ll|rrr}
\toprule
\hline
& & Table & Box & Shelf \\
\midrule
\multirow{2}{*}{RPTL (Ours)} & MFD ($\times 10^{-2}$) & $\bb{3.1 \pm 0.2}$ & $\bb{3.9 \pm 0.2}$ & $\bb{5.3 \pm 0.9}$ \\
& WD ($\times 10^{-2}$) & \textbf{2.6} & \textbf{2.9} & \textbf{3.8} \\
\midrule
\multirow{2}{*}{Linear} & MFD ($\times 10^{-2}$) & $10.6 \pm 0.3$ & $7.4 \pm 0.1$ & $9.7 \pm 1.0$ \\
& WD ($\times 10^{-2}$) & 6.9 & 6.9 & 6.3 \\
\midrule
\multirow{2}{*}{NN} & MFD ($\times 10^{-2}$) & $17.8 \pm 0.8$ & $33.2 \pm 0.6$ & $7.9 \pm 0.9$ \\
& WD ($\times 10^{-2}$) & 10.3 & 19.1 & 9.2 \\
\hline
\bottomrule
\end{tabular}
\end{adjustbox}
\end{table}

 The quantitative performance of our method, along with baseline comparisons, are given in \cref{tab:comparison}. The low distances between our learned model and the test set highlight the ability of our method to train a generative model that is consistent with the trajectories a human user would expect. This is reiterated when qualitatively examining \cref{fig:experiment_example}, where the top row consists of two images along with user-sketched trajectories at different views, and the bottom rows contain samples from the 3D trajectory model. The produced end-effector motions are consistent with the sketches provided.

\subsection{Tracing Out Letters: a Case of Intricate Trajectories}
We seek to investigate whether RPTL can be applied to learn more complex distributions of motion trajectories which may involve multiple turns or swerves. To this end, we diagrammatically teach a simulated Franka to trace out the letters ``B'', ``Z'', ``U''. This requires motion that deviates greatly from a linear trajectory and would be challenging to describe with a motion planning cost function, such as in \cite{GeoFab_gloabL_opt, PDMP}. We randomly select robot starting configurations facing a surface, and conditionally sample motion trajectories to sketch out the letters on the surfaces. These are illustrated in \cref{fig:letter_draw}. An additional tracing of the letter ``R'' is shown in \cref{fig:positions} We observe that RPTL can generate trajectories that accurately capture the intricacies of each of the characters. Additionally, we validate that RPTL can generate trajectories conditional on newly sampled end-effector positions in the vicinity. 

\subsection{Diagrammatic Teaching in the Real-world}
To test the robustness of RPTL in the real world, we diagrammatically teach a 6-DOF Unitree Z1 manipulator new skills. Additionally, we also demonstrate the applicability of RPTL to a Unitree Aliengo quadruped with the Z1 arm attached. We take two photos with a camera, find the camera poses via AprilTags \cite{AprilTag}, and collect demonstrations from the user. The robot end-effector, along with the quadruped, then tracks trajectories sampled from the trained models. 

We diagrammatically teach skills for the following tasks to the Z1 arm: \textbf{Push box}: Push a box from its side, such that it drops off the table; \textbf{Drop into cup}: Move past a mug, and hover right over a paper cup. The gripper is subsequently released and the held object shall be dropped into the paper cup; \textbf{Tip box}: Reach into a box and tip it by moving in a ``U''-shaped motion. We then mount the arm onto a quadruped and teach the quadruped + arm robot skills for the following tasks: \textbf{Drop into box}: Reach out towards a box on the floor. The gripper is released and the object held is placed into the box; \textbf{Close a drawer}: Close an open drawer.

A video recording of the robots performing each learned skill is provided, and images of the provided demonstrations and subsequent execution are given in \cref{fig:2d_to_3d,,fig:draw_close_example}. We observe that RPTL can robustly teach the robot the specified skills on both robot platforms. Moreover, Diagrammatic Teaching demonstrates its utility on mobile manipulators where kinesthetic teaching would be impractical. In particular, the task \textbf{Drop into box} requires the dog to bend its knees while the arm moves towards the box for the mounted arm to reach sufficiently low. The task \textbf{Close drawer} requires even more coordination between the dog and the arm, as the dog needs to concurrently march forward towards the drawer as the arm shuts the drawer. 

\section{Conclusions and Future Work}\label{sec:conclusions}
We introduce Diagrammatic Teaching, with which a user specifies desired motion via sketched curves, as a novel paradigm for learning from demonstration. We present Ray-tracing Probabilistic Trajectory Learning (RTPL), which estimates probability densities over 2D user sketches and traces rays into the robot's task space to fit a distribution of trajectories. We evaluate RTPL both in simulation and on real-world robots. Future avenues to explore include (1) actively generating posed images from a scene model, such that they are the most conducive for the user to sketch on; (2) incorporating user-sketched constraints, such as enforcing the elbow of a manipulator to not enter a specified region.   

\bibliographystyle{ieeetr} 
\bibliography{bib}
\end{document}